\documentclass[letterpaper]{article} % DO NOT CHANGE THIS
\usepackage[preprint]{conference}  % DO NOT CHANGE THIS
\usepackage[hyphens]{url}  % DO NOT CHANGE THIS
\usepackage{graphicx} % DO NOT CHANGE THIS
\usepackage{natbib}  % DO NOT CHANGE THIS AND DO NOT ADD ANY OPTIONS TO IT
\usepackage{caption} % DO NOT CHANGE THIS AND DO NOT ADD ANY OPTIONS TO IT
\usepackage{amsmath}
\usepackage{amssymb}
\usepackage{amsfonts}

\usepackage{algorithm}
\usepackage{algpseudocode}

\usepackage{booktabs}
\usepackage{multirow}
\usepackage{makecell}
\usepackage{array}
\usepackage{subcaption}
\usepackage{pifont}
\usepackage[table]{xcolor}

\newcommand{\cmark}{\ding{51}}
\newcommand{\xmark}{\ding{55}}
\definecolor{irislight}{HTML}{E4EAF3}

\makeatletter
\renewcommand{\@fnsymbol}[1]{%
  \ensuremath{\ifcase#1\or \dagger\or \ddagger\or \mathsection\or
  \mathparagraph\or \|\or **\or \dagger\dagger\or \ddagger\ddagger
  \else\@ctrerr\fi}}
\makeatother

\title{TASQ: Temporal-Adaptive Bit Sparsification Quantization for Diffusion Models}

\author{
    Seokho Han\textsuperscript{\rm 1},
    Dongwei Wang\textsuperscript{\rm 2},
    Jinhee Kim\textsuperscript{\rm 3},\\
    Yiran Chen\textsuperscript{\rm 3},
    Kang Eun Jeon\textsuperscript{\rm 4}\corresponding,
    Huanrui Yang\textsuperscript{\rm 2}\corresponding,
    Jong Hwan Ko\textsuperscript{\rm 1}\corresponding
}
\affiliations{
    \textsuperscript{\rm 1}Department of Electrical and Computer Engineering, Sungkyunkwan University, Korea\\
    \textsuperscript{\rm 2}Department of Electrical and Computer Engineering, University of Arizona, USA\\
    \textsuperscript{\rm 3}Department of Electrical and Computer Engineering, Duke University, USA\\
    \textsuperscript{\rm 4}Kim Jaechul Graduate School of AI, Korea Advanced Institute of Science and Technology (KAIST)\\
    \{beppa2396, jhko\}@skku.edu, kejeon@kaist.ac.kr, huanruiyang@arizona.edu
}

\begin{document}

\maketitle

\begin{abstract}
Static quantization assigns one weight precision to every denoising step. To preserve quality, that precision must accommodate the most quantization-sensitive step, even though many other steps can tolerate fewer bits. The resulting model may satisfy its memory budget, but it repeatedly pays worst-case arithmetic throughout the denoising trajectory. We introduce \textbf{Temporal-Adaptive Bit Sparsification Quantization (TASQ)} to separate these two costs. TASQ stores one shared maximum-precision weight buffer and learns a \textbf{Temporal-Spatial LSB Mask} that selects a lower effective precision for each layer and denoising stage by truncating least-significant bits. Storage therefore remains fixed by the worst case, while BitOPs decrease at less sensitive stages without per-stage weight copies or runtime search. A \textbf{Temporal-Precision Engine} maps the learned schedule to bit-serial execution, where cycles scale with effective precision and switching precision has no measured cycle overhead. On PixArt-$\Sigma$, SANA-1.6B, and SDXL-Turbo, TASQ achieves quality comparable to static quantization with less computation. Together with the Temporal-Precision Engine, it reduces execution cycles by 25--50\% over static quantization and by $6.1$--$7.5\times$ over a naive static 8-bit bit-serial execution. Code is available at \url{https://github.com/seokho-han/tasq}.
\end{abstract}

% \section{Introduction}

% Diffusion models~\citep{DiffGan,Denoisingdiff,rombach2021highresolution,DBLP:journals/cvm/WangPLGH25} generate high-quality images through an iterative denoising process, repeatedly evaluating the same network from a noisy state to a clean sample. Quantization reduces the memory and arithmetic cost of each evaluation~\citep{gholami2021survey}, and recent methods compress diffusion weights and activations to 4 bits and below. These methods nevertheless use one fixed weight precision throughout the trajectory.

% In static quantization, the required weight precision is set by the most
% quantization-sensitive timestep. If layer $l$ requires $b_{t,l}$ bits at timestep $t$, a static model must choose $b_l^{\mathrm{static}}\!\geq\!\max_t b_{t,l}$ to protect its most sensitive timestep. The same $b_l^{\mathrm{static}}$ is then used at every other timestep, including those that tolerate fewer bits. Thus the model must \emph{store} enough information for the worst case, but it need not \emph{compute} with all of that information at every step.
\section{Introduction}

% Diffusion models~\citep{DiffGan,Denoisingdiff,rombach2021highresolution,DBLP:journals/cvm/WangPLGH25}
% generate high-quality images through an iterative denoising process,
% repeatedly evaluating the same network as a noisy input is transformed
% into a clean sample. Static quantization reduces the memory and arithmetic
% cost of each evaluation~\citep{gholami2021survey}, and recent methods
% compress diffusion weights and activations to 4 bits and below.

% However, static quantization uses the same weight precision throughout
% the entire denoising trajectory. This precision must be high enough for
% the most quantization-sensitive timestep, and is therefore also used at
% less sensitive timesteps that could tolerate fewer bits. As a result, the
% model must store enough information for the worst case, but does not need
% to compute with all of that information at every step.
Diffusion models~\citep{DiffGan,Denoisingdiff,rombach2021highresolution,DBLP:journals/cvm/WangPLGH25}
generate high-quality images through an iterative denoising process,
repeatedly inferring the same network to transform a noisy input into a clean sample. Quantization reduces the memory and arithmetic cost of each inference~\citep{gholami2021survey} by compressing diffusion weights and activations to 4 bits and lower~\cite{li2023qdiffusion,chen2024QDiT}. 

\begin{figure}[t]
    \centering
    \includegraphics[width=0.95\linewidth]{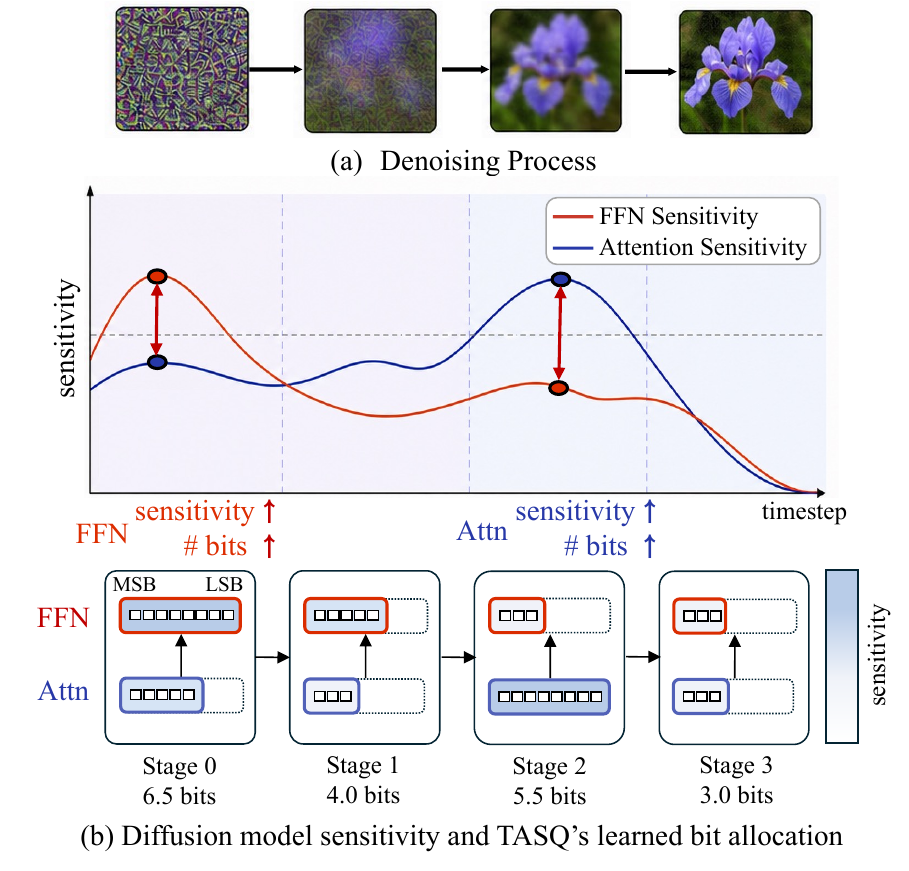}
    \vspace{-12pt}
    \caption{Temporal sensitivity and TASQ bit allocation across diffusion stages: FFN sensitivity peaks at noisy steps while attention peaks mid-trajectory, and TASQ assigns higher bits where sensitivity is high.}
    \label{fig:intro_sensitivity}
\end{figure}

Meanwhile, we notice a redundancy in Diffusion computation that existing quantization methods fail to address. As shown in Figure~\ref{fig:intro_sensitivity}, quantization sensitivity varies across both timesteps and layer types. FFN layers are most sensitive near the noisy end, whereas attention layers peak later. 
While mixed-precision quantization method explores layer-wise sensitivity through search-based~\citep{wang2019haq,dong2019hawq} or learning-based~\citep{yang2021bsq,csq,seokho2025msq} methods, they maintain same precision across the timesteps, failing to exploit the temporal variation. 
% While some timestep-aware diffusion quantizers exist~\citep{MPQdiff,MPQdm,TCAQdm,AdaTSQ}, the exploration stops at changing calibration parameters with precision unchanged.
Recent works adapt precision over denoising steps, but only for activations~\citep{AdaTSQ}. This is limiting because activation quantization degrades sharply below 3 bits due to outliers~\citep{li2024svdquant}, while weight precision remains fixed at the worst-case timestep.

Under such static quantization, the model has to take a precision that stores enough information for the worst case of each layer across all timesteps; while wasting computation in the steps that are less sensitive.

To reduce the temporal redundancy in Diffusion computation, we propose TASQ, the first temporal-adaptive quantizer that can assign different weight precision to model layers at different diffusion steps. Changing precision across timesteps without memory overhead requires TASQ to learn multiple step-specific operating precisions from one shared weight representation, rather than storing the weight of each step separately.
% TASQ avoids those copies by nesting all operating precisions in one buffer. Under a truncation-consistent quantizer, the $b$-bit representation is a prefix of the stored $b_{\max}$-bit code. TASQ learns, for every layer and temporal stage, how many least-significant planes can be omitted. The resulting \textbf{Temporal-Spatial LSB Mask} changes effective precision without changing the stored weights, and \textbf{Farthest-Stage-First Training} stabilizes the joint optimization of shared weight bits and stage-specific masks. At inference, the learned allocation is a table lookup followed by bit-plane truncation; there is no per-step search, model reload, or weight repacking.
TASQ therefore takes a truncation-based method. From a shared high-precision weight, a \textbf{Temporal-Spatial LSB Mask} is proposed to determine how
many least-significant bits can be omitted for each layer and stage.
To avoid gradient complication for jointly training multiple-precision weights across timesteps, we further propose \textbf{Farthest-Stage-First Training} that
separates the weight updates across distant stages. At inference, TASQ requires
only a table lookup for precision determination and bit-plane truncation for operating weight generation, without additional weight
copies, search, or repacking.
Our main contributions are summarized as follows:
% \begin{itemize}
%     % \item We distinguish storage precision from operating precision in
%     % diffusion quantization: the former covers the most sensitive timestep,
%     % while the latter can vary by layer and stage to reduce BitOPs.
%     \item Unlike prior methods that use a single weight precision throughout the denoising trajectory, we decouple storage precision from operating precision, allowing the latter to vary across layers and stages to reduce BitOPs.
%     \item We formulate temporal weight allocation over a single shared
%     weight representation, using Temporal-Spatial LSB Masking to learn
%     stage- and layer-wise operating precisions.
%     % \item We implement the Temporal-Precision Engine, achieving
%     % $33$--$50\%$ fewer cycles than static quantization at matched quality
%     % and $3.2$--$5.1\times$ fewer cycles than a naive bit-plane loop.
%     \item We implement the Temporal-Precision Engine, a bit-serial accelerator that streams only the required weight planes. Combined with TASQ's learned precision schedule, it achieves $6.1$--$7.5\times$ fewer cycles than static 8-bit execution while preserving FID and ImageReward.
% \end{itemize}
\begin{itemize}
    \item We decouple storage and operating precision, allowing weight precision to vary across layers and denoising stages to reduce BitOPs.
    \item We propose Temporal-Spatial LSB Mask, which learns stage and layer-wise precision over a single shared weight representation.
    \item We design the Temporal-Precision Engine, which streams only the selected weight planes and achieves $6.1$--$7.5\times$ fewer cycles than a naive static 8-bit execution while preserving FID and ImageReward.
\end{itemize}

\section{Related Works}

\noindent\textbf{Diffusion Model Quantization.}
Diffusion quantization first established 8-bit PTQ baselines with Q-Diffusion~\citep{li2023qdiffusion} and PTQ4DM~\citep{shang2023ptqdm}. Later work explored sensitivity-aware calibration~\citep{yang2023efficient}, timestep-aware quantization~\citep{huang2024tfmqdm,he2023ptqd,wang2024towards}, text-to-image models~\citep{tang2024progressive}, and DiT backbones~\citep{wu2024ptq4dit,chen2024QDiT}. LoRA-based QAT methods such as EfficientDM~\citep{hu2022lora,he2024efficientdm} recover quality with limited retraining, while BinaryDM~\citep{zheng2024binarydm}, QuEST~\citep{wang2024quest}, and BitsFusion~\citep{sui2024bitsfusion} target sub-2-bit weights. These methods allocate precision spatially and reuse that allocation across timesteps.

Several low-bit pipelines protect selected components---error-sensitive tokens in ViDiT-Q~\citep{zhao2024viditq}, the BOS text token in MixDQ~\citep{zhao2024mixdq}, or a 16-bit low-rank branch in SVDQuant~\citep{li2024svdquant}---and quantize the remaining model more aggressively. The protected components and bit allocation are fixed after calibration. The most sensitive timestep therefore sets a precision floor for the rest of the trajectory.

\noindent\textbf{Mixed-Precision and Bit-Level Quantization.}
Mixed-precision quantization exploits non-uniform layer-wise sensitivity. HAQ searches for layer bit-widths with reinforcement learning~\citep{wang2019haq}, while HAWQ and HAWQ-V2 use second-order sensitivity~\citep{dong2019hawq,dong2020hawqv2}; both face a combinatorial allocation space. Bit-level methods learn the allocation during training instead: BSQ optimizes individual bits~\citep{yang2021bsq}, CSQ uses a continuous relaxation~\citep{csq}, and 
%MSQ regularizes LSBs without bit-splitting overhead~\citep{seokho2025msq}.
MSQ directly regularizes the LSB itself of the quantized weights without bit-splitting overhead~\citep{seokho2025msq}.

%These methods nevertheless remain spatial: one learned allocation is reused at every timestep.
These methods nevertheless learn one fixed spatial weight-precision allocation. In contrast, TASQ learns multiple stage-specific operating precisions by applying different masks to one shared quantized weight.

\noindent\textbf{Timestep-Aware Quantization.}
A recent line of work makes diffusion quantization timestep-aware, but not through dynamic weight precision. TCAQ-DM~\citep{TCAQdm} adapts activation ranges at a uniform bit-width; MPQ-DM~\citep{MPQdm} and MPQ-Diff~\citep{MPQdiff} choose per-layer allocations that remain fixed across steps; and AdaTSQ~\citep{AdaTSQ} searches per-timestep activation bits while keeping weights static to avoid memory overhead. These methods are complementary to TASQ, which learns the effective \emph{weight} precision for each layer and stage from a single shared buffer. Table~\ref{tab:positioning} summarizes this distinction.

\section{Methods}

\begin{figure*}[t]
    \centering
    \includegraphics[width=0.9\textwidth]{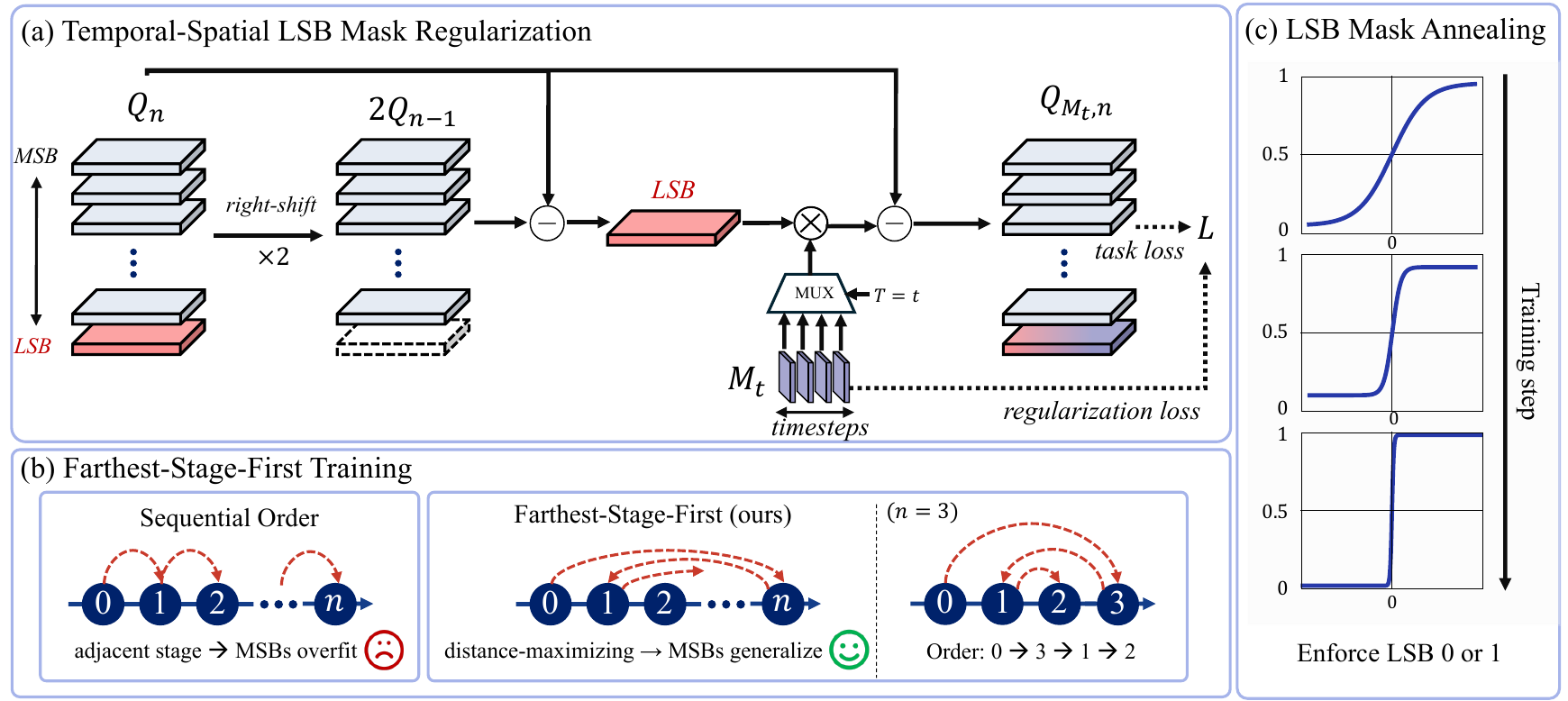}
    \vspace{-6pt}
    \caption{Process of Temporal--Spatial LSB Mask Regularization and LSB Mask Annealing.}
    \label{fig:mask_process}
\end{figure*}
We propose TASQ, a temporal-adaptive quantization framework that varies weight precision across denoising stages. TASQ uses a shared weight representation that allows lower-bit weights to be obtained by direct truncation. Because precision choices are coupled through the shared weights, we introduce Temporal-Spatial LSB Masking to learn the precision allocation, and optimize it together with
the shared weights using parameter-efficient LoRA-based QAT. However, jointly updating the shared weights across neighboring stages
can produce highly correlated gradients and cause training to plateau. To address this issue, Farthest-Stage-First
Training separates the updates across distant denoising stages and stabilizes
optimization. Since operating precision changes across the denoising
trajectory, we also co-design the Temporal-Precision Engine to execute the
resulting precision schedule efficiently by reusing activation bit planes
and streaming only the required weight bits.
\subsection{Preliminaries}
\label{sec:method-prelim}

\paragraph{Quantizer for Adaptive Precision.}
% To develop a single model that supports various integer precisions, recent studies have proposed nesting smaller bit-widths within larger ones. The simplest implementation is via bit-shifting, which mandates strict alignment between truncation and quantization~\citep{jin2020adabitsneuralnetworkquantization,kim2025truncquant}. For a normalized weight $W \in [0,1]$, the $n$-bit quantized value and its directly truncated $(n-1)$-bit counterpart are jointly defined as:
% \begin{equation}
% Q_n = \left\lfloor W \cdot 2^n \right\rfloor, \qquad Q_{n-1} = \left\lfloor \frac{Q_n}{2} \right\rfloor.
% \label{eq:truncquant}
% \end{equation}
% where $\lfloor \cdot \rfloor$ denotes the floor operation. Equation~\eqref{eq:truncquant} makes the lower-bit code a prefix of the higher-bit code. Switching precision therefore requires only truncation or a right shift, rather than a separately quantized weight copy.
For a normalized weight $W \in [0,1]$, the $n$-bit quantized weight and
its directly truncated $(n-1)$-bit weight are defined as
\begin{equation}
Q_n = \left\lfloor W \cdot 2^n \right\rfloor, \qquad
Q_{n-1} = \left\lfloor \frac{Q_n}{2} \right\rfloor.
\label{eq:truncquant}
\end{equation}
Here, $\lfloor \cdot \rfloor$ denotes the floor operation.
Equation~\eqref{eq:truncquant} ensures that the lower-precision weight is
obtained by removing the least-significant bit from the higher-precision
weight. Switching precision therefore requires only truncation or a right
shift, rather than a separately quantized weight copy.

\paragraph{Efficient QAT using LoRA.}
To bridge the gap between the performance of Quantization-Aware Training (QAT) and the efficiency of Post-Training Quantization (PTQ), recent studies~\citep{he2024efficientdm,jeon2025l4qparameterefficientquantizationaware} propose integrating LoRA into the quantization. Specifically, these methods apply the quantization operator to the effective weight formed by merging learnable low-rank adapters with frozen original weights, formulated as:
\begin{equation}
Y = Q(X)Q(W + BA),
\label{eq:qalora}
\end{equation}
where $W$ represents the frozen weights, and $B$ and $A$ denote the trainable low-rank matrices. This yields fully quantized weights for efficient bit-wise inference while training only the low-rank parameters.

\subsection{TASQ}
TASQ starts from a single quantized weight representation for each layer and introduces learnable Temporal-Spatial LSB masks to decompose it into multiple stage-specific operating precisions. This allows the model to vary its operating precision without storing separate weight copies. Unlike previous bit-level methods that directly regularize the bits of quantized weights, TASQ optimizes the masks jointly with the LoRA parameters. Activation precision remains fixed in this work and can be adapted independently.

% \paragraph{Storage precision.} The deployment memory budget determines $b_l^{\max}$, which must cover the most sensitive stage. Our main evaluation uses a shared $b_{\max}{=}8$ buffer to expose the full quality--compute frontier; the same construction applies when a tighter deployment budget sets a smaller $b_{\max}$ (Supplementary Section~C).

\subsubsection{Temporal-Spatial LSB Mask}
For a stored $n$-bit quantized weight, we extract its least-significant bit
from the difference between $Q_n$ and the truncated $(n-1)$-bit weight:
\begin{equation}
\mathrm{LSB} = Q_n - 2 \cdot Q_{n-1}.
\label{eq:lsb_extraction}
\end{equation}
Although the LSB is extracted from the shared quantized weight, it remains
shared across all stages and layers. We therefore introduce a mask
$M_{t,l}$ that allows the same LSB to take different sparsity states across
stages and layers:
\begin{equation}
Q_{M,n}^{(t,l)} = Q_n - (1 - M_{t,l}) \cdot \mathrm{LSB}.
\label{eq:lsb_masking}
\end{equation}
Here, $M_{t,l}{=}1$ retains the bit, while $M_{t,l}{=}0$ removes it.
Applying the same operation recursively to the remaining least-significant
bits determines the operating precision while preserving the shared
high-precision weight.

\paragraph{Training Objective.}
We jointly optimize the LoRA parameters and masks using the full-precision model as a teacher:
\begin{equation}
\mathcal{L}_{\mathrm{total}} = \lVert \epsilon_q - \epsilon_{\mathrm{fp}} \rVert_2^2 + \lambda_M \sum_{t,l} M_{t,l}.
\label{eq:final_objective}
\end{equation}
The first term matches the predicted noise, while the second penalizes
the retained least-significant bits.
Under the straight-through estimator, $Q_n$ and $Q_{n-1}$ share the same weight-gradient path up to their power-of-two scale, so the extracted LSB is treated as locally constant:
\begin{equation}
\frac{\partial \mathrm{LSB}}{\partial W} \approx 0.
\label{eq:lsb_zero_grad}
\end{equation}
The weight gradient therefore follows the masked quantized output:
\begin{equation}
\frac{\partial \mathcal{L}}{\partial W} \approx \frac{\partial \mathcal{L}}{\partial Q_{M,n}^{(t,l)}}.
\label{eq:l_r_approx}
\end{equation}
Differentiating Eq.~\eqref{eq:lsb_masking} with respect to $M_{t,l}$ gives
\begin{equation}
\frac{\partial \mathcal{L}}{\partial M_{t,l}} = \frac{\partial \mathcal{L}}{\partial Q_{M,n}^{(t,l)}} \cdot \mathrm{LSB} + \lambda_M,
\label{eq:mask_grad}
\end{equation}
Thus, the LSB is retained when its contribution to the distillation loss outweighs the regularization $\lambda_M$; otherwise, the mask is driven toward zero. Figure~\ref{fig:mask_process} illustrates this process.

\begin{figure*}[t]
    \centering
    \includegraphics[width=0.98\textwidth]{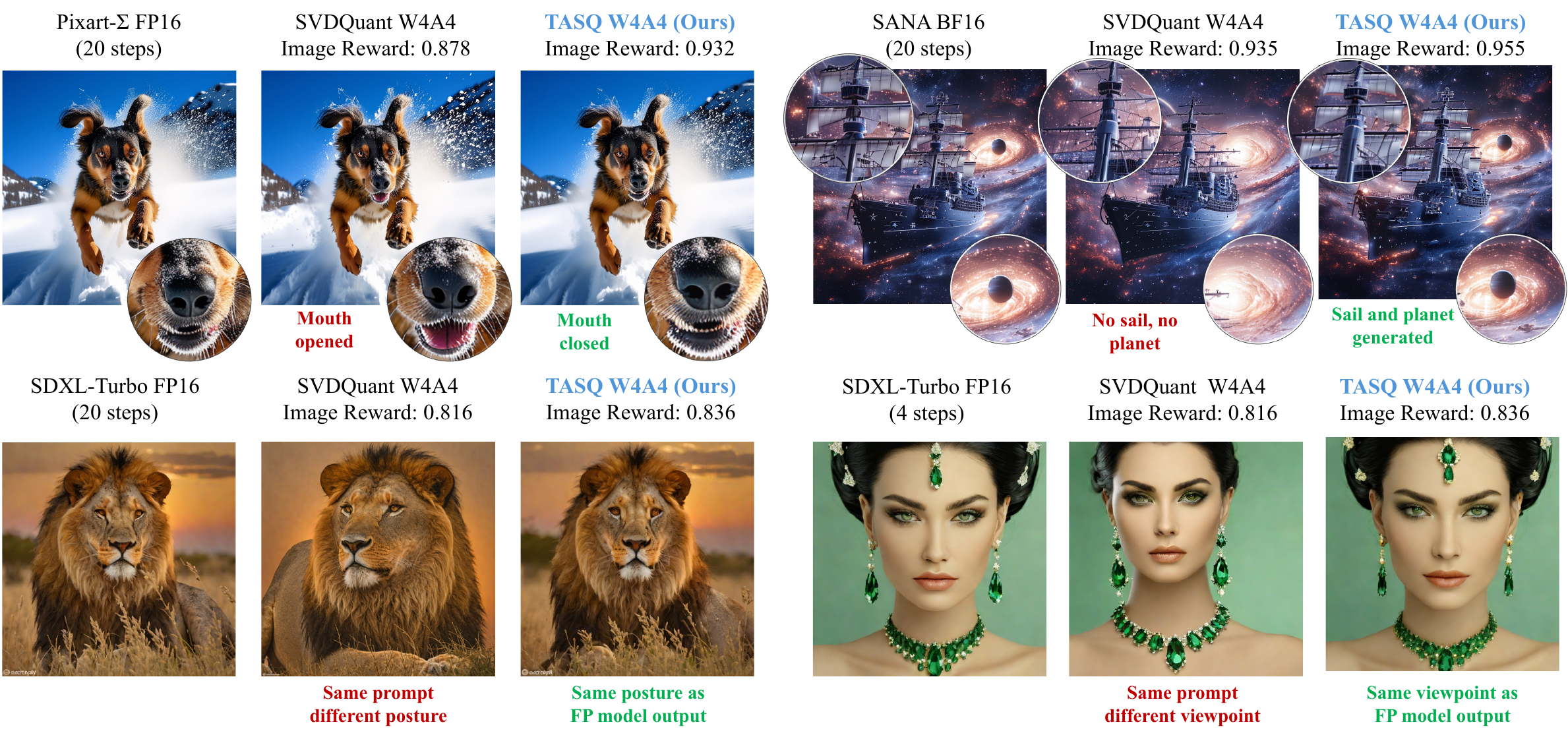}
    \vspace{-10pt}
    \caption{Qualitative Image Generation Results on PixArt-$\Sigma$, SANA-1.6B, SDXL-Turbo.}
    \label{fig:sample_image}
\end{figure*}

\paragraph{LSB Mask Annealing.}
Direct binary selection is not differentiable, so we use a continuous gate during training and anneal it to a binary decision, following the sparsification schedule of PAT~\citep{liu2025pat}. For training step $s$ and annealing horizon $s_0$, define
\begin{equation}
\resizebox{\columnwidth}{!}{$\displaystyle
\tau(s) =
\begin{cases}
\dfrac{1}{1 - \dfrac{\ln(s)}{\ln(s_0)}}, & s < s_0, \\[2ex]
\epsilon^{-1}, & \text{otherwise}
\end{cases}
\qquad
\beta(s) =
\begin{cases}
-\dfrac{s}{s_0} + 0.5, & s < s_0/2, \\[2ex]
0, & \text{otherwise}
\end{cases}
$}
\label{eq:tau_beta_def}
\end{equation}
The continuous gate is
\begin{equation}
M_{t,l} = \mathcal{G}(s,m_{t,l}) = \frac{1}{1 + e^{-\tau(s)\cdot m_{t,l}}} + \beta(s),
\label{eq:mask_function}
\end{equation}
where $m_{t,l}$ is learned. The bias initializes the gate near one, and increasing temperature binarizes it.

% \subsubsection{Progressive Training over Stages}
% \label{sec:method-progressive}
\paragraph{Stage-wise Training.}
Adjacent timesteps learn similar bit allocations, consistent with the temporal redundancy observed in diffusion trajectories~\citep{whalen2025earlybirddiffusion,shi2025closer}. We therefore partition the trajectory into contiguous stages and share $M_{t,l}$ within each stage. This reduces the number of mask parameters while retaining the temporal variation; the elbow falls at four stages (Appendix~\ref{sec:appendix-number-of-stages}).

\paragraph{Farthest-Stage-First Training.}
Because all stages share the same MSBs, training neighboring stages in succession leads to highly correlated updates and caused the loss to plateau early in our experiments. We therefore train stages in a farthest-stage-first order, placing consecutive updates as far apart as possible along the trajectory. For four stages, the order is $0\!\to\!3\!\to\!1\!\to\!2$. This exposes the shared MSBs to more diverse parts of the trajectory while preserving a separate LSB mask for each stage.
\begin{table}[t]
\centering
\scriptsize
\setlength{\tabcolsep}{4pt}
\caption{Bit-plane dataflows for $L$ weight and $R{=}4$ activation planes. Keeping activations on chip reduces the fetch count to $L{+}R$, bit-exactly. Algorithm~\ref{alg:engine} gives the schedules.}
\vspace{-9pt}
\label{tab:codesign}
\begin{tabular}{lccccc}
\toprule
& \multicolumn{2}{c}{Fetch planes} & \multicolumn{3}{c}{Total cycles} \\
\cmidrule(lr){2-3}\cmidrule(lr){4-6}
Schedule & model & $L{=}8$ & $L{=}3$ & $L{=}4$ & $L{=}8$ \\
\midrule
Naive (per-pass refetch) & $2LR$ & 64 & 13{,}853 & 18{,}413 & 36{,}653 \\
Weight-stationary        & $L{+}LR$ & 40 & 8{,}723 & 11{,}573 & 22{,}973 \\
\rowcolor{irislight}\textbf{Activation-stationary (ours)} & $\mathbf{L{+}R}$ & \textbf{12} & \textbf{4{,}313} & \textbf{4{,}883} & \textbf{7{,}163} \\
\bottomrule
\end{tabular}
\end{table}
\begin{table*}[t!]
\setlength{\tabcolsep}{3.4pt}
\renewcommand{\arraystretch}{0.94}
\caption{
Quality--compute comparison. ``Eff. W'' denotes the average operating weight precision and ``Rel. BitOPs'' the compute cost normalized to static W8A8. TASQ stores one shared 8-bit weight buffer across all stages. W4 rows are compute-matched, while W8 TASQ retains W8 quality with 74--75\% compute. Non-SVDQuant controls are reported in Table~\ref{tab:no_svdquant}.
}
\vspace{-10pt}

\label{tab:quality}
\scriptsize
\centering
\resizebox{0.96\textwidth}{!}{%
\begin{tabular}{cccccccccccccc}
\toprule
& & & & & & \multicolumn{4}{c}{MJHQ} & \multicolumn{4}{c}{sDCI} \\
\cmidrule(lr){7-10} \cmidrule(lr){11-14}
Backbone & Model & A-Bit & \makecell{Eff.\\W-Bit} & \makecell{Rel.\\BitOPs} & Method &
FID ($\downarrow$) & IR ($\uparrow$) & LPIPS ($\downarrow$) & PSNR ($\uparrow$) &
FID ($\downarrow$) & IR ($\uparrow$) & LPIPS ($\downarrow$) & PSNR ($\uparrow$) \\
\midrule

% =====================  PixArt-Sigma  =======================
\multirow{22}{*}{DiT}
& \multirow{12}{*}{\makecell{PixArt-$\Sigma$\\(20 Steps)}}
& 16 & 16 & 4.00 & FP & 16.6 & 0.944 & -- & -- & 24.8 & 0.966 & -- & -- \\
\cmidrule(lr){3-14}

% --- A8, W8 ---
& & \multirow{3}{*}{8}
    & 8.00 & 1.00 & ViDiT-Q          & 15.7 & 0.944 & 0.137 & 22.5 & \textbf{23.5} & \textbf{0.974} & 0.163 & 20.4 \\
& & & 8.00 & 1.00 & SVDQuant         & 16.3 & \textbf{0.955} & \textbf{0.109} & \textbf{23.7} & 24.2 & 0.969 & \textbf{0.129} & \textbf{21.8} \\
& & & \cellcolor{irislight}\textbf{5.91} & \cellcolor{irislight}\textbf{0.74} & \cellcolor{irislight}\textbf{SVDQuant+TASQ}
    & \cellcolor{irislight}\textbf{15.6} & \cellcolor{irislight}\textbf{0.955} & \cellcolor{irislight}0.112 & \cellcolor{irislight}\textbf{23.7}
    & \cellcolor{irislight}23.9 & \cellcolor{irislight}0.968 & \cellcolor{irislight}0.132 & \cellcolor{irislight}\textbf{21.8} \\
\cmidrule(lr){3-14}

% --- A8, W4 ---
& & \multirow{4}{*}{8}
    & 4.00 & 0.50 & ViDiT-Q          & 37.3 & 0.573 & 0.611 & 12.0 & 40.6 & 0.600 & 0.629 & 11.2 \\
& & & 4.00 & 0.50 & SVDQuant         & 17.8 & 0.915 & 0.290 & 19.2 & 24.6 & 0.942 & 0.315 & 17.8 \\
& & & 4.00 & 0.50 & SVDQuant+QAT     & 17.0 & 0.928 & 0.268 & 19.8 & 24.2 & 0.955 & 0.198 & 18.5 \\
& & & \cellcolor{irislight}\textbf{4.00} & \cellcolor{irislight}\textbf{0.50} & \cellcolor{irislight}\textbf{SVDQuant+TASQ}
    & \cellcolor{irislight}\textbf{16.5} & \cellcolor{irislight}\textbf{0.942} & \cellcolor{irislight}\textbf{0.242} & \cellcolor{irislight}\textbf{20.5}
    & \cellcolor{irislight}\textbf{23.8} & \cellcolor{irislight}\textbf{0.968} & \cellcolor{irislight}\textbf{0.173} & \cellcolor{irislight}\textbf{19.2} \\
\cmidrule(lr){3-14}

% --- A4, W4 ---
& & \multirow{4}{*}{4}
    & 4.00 & 0.25 & ViDiT-Q          & 412 & -2.27 & 0.854 & 6.44 & 425 & -2.28 & 0.838 & 6.70 \\
& & & 4.00 & 0.25 & SVDQuant         & 19.2 & 0.878 & 0.323 & 17.6 & 25.9 & 0.918 & 0.352 & 16.5 \\
& & & 4.00 & 0.25 & SVDQuant+QAT     & 17.8 & 0.901 & 0.308 & 18.0 & 24.8 & 0.938 & 0.318 & 17.3 \\
& & & \cellcolor{irislight}\textbf{3.98} & \cellcolor{irislight}\textbf{0.25} & \cellcolor{irislight}\textbf{SVDQuant+TASQ}
    & \cellcolor{irislight}\textbf{16.1} & \cellcolor{irislight}\textbf{0.932} & \cellcolor{irislight}\textbf{0.282} & \cellcolor{irislight}\textbf{18.6}
    & \cellcolor{irislight}\textbf{23.1} & \cellcolor{irislight}\textbf{0.966} & \cellcolor{irislight}\textbf{0.263} & \cellcolor{irislight}\textbf{18.8} \\

\cmidrule(lr){2-14}

% =====================  SANA-1.6B  ==========================
& \multirow{10}{*}{\makecell{SANA\\-1.6B\\(20 Steps)}}
& 16 & 16 & 4.00 & FP & 16.2 & 1.10 & -- & -- & 22.4 & 1.07 & -- & -- \\
\cmidrule(lr){3-14}

% --- A8, W8 ---
& & \multirow{2}{*}{8}
    & 8.00 & 1.00 & SVDQuant         & \textbf{15.9} & 1.095 & \textbf{0.243} & \textbf{18.4} & 22.4 & 1.016 & \textbf{0.221} & \textbf{17.5} \\
& & & \cellcolor{irislight}\textbf{5.98} & \cellcolor{irislight}\textbf{0.75} & \cellcolor{irislight}\textbf{SVDQuant+TASQ}
    & \cellcolor{irislight}\textbf{15.9} & \cellcolor{irislight}\textbf{1.096} & \cellcolor{irislight}0.274 & \cellcolor{irislight}17.6
    & \cellcolor{irislight}\textbf{22.2} & \cellcolor{irislight}\textbf{1.066} & \cellcolor{irislight}0.291 & \cellcolor{irislight}16.3 \\
\cmidrule(lr){3-14}

% --- A8, W4 ---
& & \multirow{3}{*}{8}
    & 4.00 & 0.50 & SVDQuant         & 17.7 & 1.018 & 0.241 & 17.7 & 22.5 & 1.018 & \textbf{0.264} & 16.3 \\
& & & 4.00 & 0.50 & SVDQuant+QAT     & 16.8 & 1.048 & 0.252 & 17.7 & 22.5 & 1.032 & 0.271 & 16.4 \\
& & & \cellcolor{irislight}\textbf{3.99} & \cellcolor{irislight}\textbf{0.50} & \cellcolor{irislight}\textbf{SVDQuant+TASQ}
    & \cellcolor{irislight}\textbf{15.9} & \cellcolor{irislight}\textbf{1.096} & \cellcolor{irislight}0.264 & \cellcolor{irislight}17.8
    & \cellcolor{irislight}\textbf{22.4} & \cellcolor{irislight}\textbf{1.060} & \cellcolor{irislight}0.280 & \cellcolor{irislight}\textbf{16.6} \\
\cmidrule(lr){3-14}

% --- A4, W4 ---
& & \multirow{4}{*}{4}
    & 4.00 & 0.25 & RTN              & 20.5 & 0.894 & 0.339 & 15.3 & 28.6 & 0.807 & 0.371 & 13.8 \\
& & & 4.00 & 0.25 & SVDQuant         & 19.3 & 0.935 & \textbf{0.220} & 17.8 & 28.1 & 0.846 & \textbf{0.242} & \textbf{16.2} \\
& & & 4.00 & 0.25 & SVDQuant+QAT     & 18.5 & 0.952 & 0.248 & 17.5 & 26.4 & 0.878 & 0.263 & 15.9 \\
& & & \cellcolor{irislight}\textbf{3.99} & \cellcolor{irislight}\textbf{0.25} & \cellcolor{irislight}\textbf{SVDQuant+TASQ}
    & \cellcolor{irislight}\textbf{16.6} & \cellcolor{irislight}\textbf{1.074} & \cellcolor{irislight}0.297 & \cellcolor{irislight}17.1
    & \cellcolor{irislight}\textbf{21.7} & \cellcolor{irislight}\textbf{1.060} & \cellcolor{irislight}0.320 & \cellcolor{irislight}15.8 \\

\midrule

% =====================  SDXL-Turbo  =========================
\multirow{12}{*}{UNet}
& \multirow{12}{*}{\makecell{SDXL-Turbo\\(4 Steps)}}
& 16 & 16 & 4.00 & FP & 24.3 & 0.845 & -- & -- & 24.7 & 0.847 & -- & -- \\
\cmidrule(lr){3-14}

% --- A8, W8 ---
& & \multirow{3}{*}{8}
    & 8.00 & 1.00 & MixDQ            & 24.1 & 0.834 & 0.147 & 21.7 & 25.0 & 0.690 & 0.157 & 21.6 \\
& & & 8.00 & 1.00 & SVDQuant         & 24.3 & \textbf{0.845} & \textbf{0.100} & 24.0 & 24.8 & 0.701 & \textbf{0.110} & \textbf{23.7} \\
& & & \cellcolor{irislight}\textbf{5.95} & \cellcolor{irislight}\textbf{0.74} & \cellcolor{irislight}\textbf{SVDQuant+TASQ}
    & \cellcolor{irislight}\textbf{24.0} & \cellcolor{irislight}\textbf{0.845} & \cellcolor{irislight}0.122 & \cellcolor{irislight}\textbf{24.2}
    & \cellcolor{irislight}\textbf{24.5} & \cellcolor{irislight}\textbf{0.730} & \cellcolor{irislight}0.121 & \cellcolor{irislight}23.5 \\
\cmidrule(lr){3-14}

% --- A8, W4 ---
& & \multirow{4}{*}{8}
    & 4.00 & 0.50 & MixDQ            & 27.7 & 0.708 & 0.402 & 15.7 & 25.9 & 0.610 & 0.415 & 15.7 \\
& & & 4.00 & 0.50 & SVDQuant         & 24.5 & 0.835 & 0.225 & 19.0 & 25.1 & 0.692 & 0.232 & 19.1 \\
& & & 4.00 & 0.50 & SVDQuant+QAT     & 24.2 & 0.840 & 0.215 & 19.4 & 24.9 & 0.698 & 0.220 & 19.4 \\
& & & \cellcolor{irislight}\textbf{3.97} & \cellcolor{irislight}\textbf{0.50} & \cellcolor{irislight}\textbf{SVDQuant+TASQ}
    & \cellcolor{irislight}\textbf{23.8} & \cellcolor{irislight}\textbf{0.845} & \cellcolor{irislight}\textbf{0.198} & \cellcolor{irislight}\textbf{19.76}
    & \cellcolor{irislight}\textbf{24.7} & \cellcolor{irislight}\textbf{0.703} & \cellcolor{irislight}\textbf{0.205} & \cellcolor{irislight}\textbf{19.77} \\
\cmidrule(lr){3-14}

% --- A4, W4 ---
& & \multirow{4}{*}{4}
    & 4.00 & 0.25 & MixDQ            & 353 & -2.26 & 0.685 & 11.0 & 373 & -2.287 & 0.686 & 11.37 \\
& & & 4.00 & 0.25 & SVDQuant         & 24.6 & 0.816 & 0.262 & 18.11 & 25.2 & 0.671 & 0.272 & 18.0 \\
& & & 4.00 & 0.25 & SVDQuant+QAT     & 24.2 & 0.823 & 0.255 & 18.25 & 25.0 & 0.678 & 0.264 & 18.15 \\
& & & \cellcolor{irislight}\textbf{3.98} & \cellcolor{irislight}\textbf{0.25} & \cellcolor{irislight}\textbf{SVDQuant+TASQ}
    & \cellcolor{irislight}\textbf{23.6} & \cellcolor{irislight}\textbf{0.836} & \cellcolor{irislight}\textbf{0.244} & \cellcolor{irislight}\textbf{18.50}
    & \cellcolor{irislight}\textbf{24.7} & \cellcolor{irislight}\textbf{0.691} & \cellcolor{irislight}\textbf{0.252} & \cellcolor{irislight}\textbf{18.45} \\

\bottomrule
\end{tabular}}
\end{table*}

\begin{figure*}[t]
    \centering

    \includegraphics[width=0.92\textwidth]{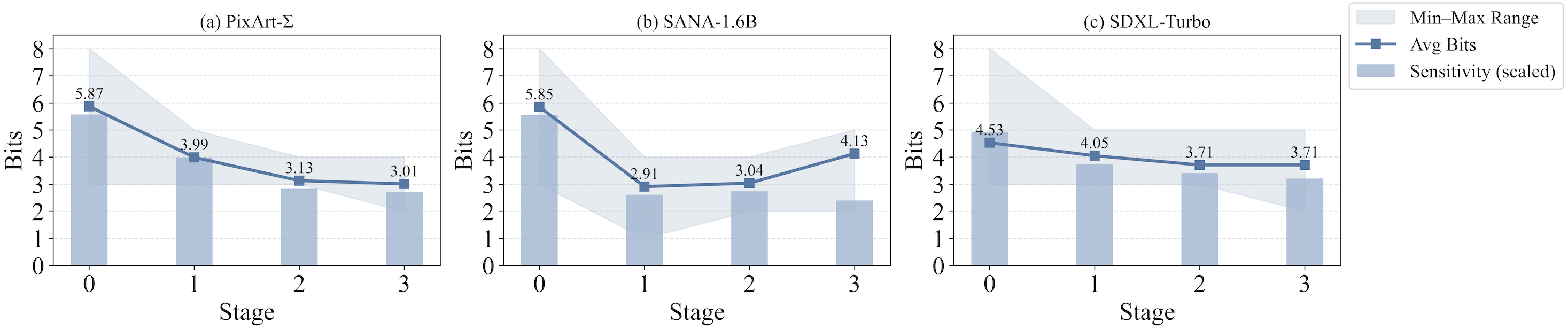}
    \vspace{-10pt}
    \caption{Effective weight precision across temporal stages under the W4A4 operating budget. The static allocation must cover the most sensitive stage, whereas TASQ follows the stage-dependent sensitivity and uses fewer bits elsewhere.}
    \label{fig:avg}
\end{figure*}

% \subsection{Algorithm--Hardware Co-design}
% \label{sec:method-codesign}
% A bit-serial GEMM decomposes an $L$-bit$\times$$R$-bit product into $L\!\cdot\!R$ binary passes. A naive loop reloads an operand on every pass. TASQ varies only the weight precision, so the \textbf{Temporal-Precision Engine} holds the $R$ activation planes on chip and streams the $L$ weight planes once, reducing the fetch count from $2LR$ to $L{+}R$. We implement this schedule in software on the unmodified BISMO accelerator~\citep{bismo}. Across the precisions in Table~\ref{tab:codesign}, it uses $3.2$--$5.1\times$ fewer cycles than the naive loop, remains bit-exact, and adds no measured cycles when the precision changes. These measurements use a fetch-bound single tile; Supplementary Section~G reports the compute-bound network setting.

\subsection{Algorithm--Hardware Co-design}
\label{sec:method-codesign}
TASQ requires hardware whose execution cost scales with the selected precision. We therefore co-design the \textbf{Temporal-Precision Engine} to execute the learned weight schedule. In bit-serial GEMM, an $L$-bit weight and an $R$-bit activation require $L \times R$ binary operations. A naive implementation reloads an operand for each pass, which limits the benefit of reducing weight precision.

Since TASQ varies only the weight precision, our engine keeps the activation bit planes on chip and streams only the weight planes selected for the current layer and stage. This reduces the number of fetched planes from $2LR$ to $L+R$ while preserving bit-exact outputs. We implement this dataflow in software based on BISMO accelerator~\citep{bismo}. As shown in Table~\ref{tab:codesign}, it reduces execution cycles by $3.2$--$5.1\times$ over a naive bit-plane loop and introduces no measured cycle overhead when precision changes.

\section{Experiments}

We report generation quality and BitOPs in Table~\ref{tab:quality}, inspect the learned allocation in Figure~\ref{fig:avg}, ablate the mask and training schedule, and measure hardware cost in \S\ref{sec:efficiency}.
To separate the effect of temporal adaptivity from that of SVDQuant
initialization, we evaluate TASQ under two settings.
Table~\ref{tab:quality} compares the original SVDQuant model, the same
model further adapted through QAT, and the model trained with TASQ.
Table~\ref{tab:no_svdquant} then compares QAT and TASQ when both are
trained without SVDQuant initialization. Together, these experiments
evaluate whether TASQ remains effective both with and without a strong
PTQ starting point.

\subsection{Experimental Setup}

\paragraph{Models and Denoising Schedules.}
We evaluate the DiT models PixArt-$\Sigma$ (0.6B)~\citep{chen2024pixartsigma} and SANA-1.6B~\citep{xie2024sana}, and the U-Net model SDXL-Turbo (2.6B)~\citep{Podell2023SDXLIL, sauer2023add}. PixArt-$\Sigma$ and SANA use 20 denoising steps grouped into four stages. SDXL-Turbo uses four steps, so each step has its own stage.

\paragraph{Quantization and Methods.}
We report W8A8, W4A8, and W4A4; Appendix~\ref{sec:appendix-quantization} gives the quantization details. The PTQ baselines are ViDiT-Q, MixDQ, RTN, and SVDQuant. The fine-tuned baselines are QAT and SVDQuant+QAT, with SVDQuant+QAT serving as the non-temporal control for SVDQuant+TASQ. SVDQuant uses a rank-16 low-rank branch at the average 6-bit setting and rank 32 at 4 bits. TASQ uses rank-32 LoRA adapters $B,A$ in all settings. Appendix~\ref{sec:appendix-lsb-mask-robustness} compares the LSB mask with a standard sigmoid gate.

\paragraph{Training.}
We train TASQ under two settings, using either pure LoRA-QAT for 5.0k iterations or SVDQuant-initialized adaptation for 1.4k iterations. In both settings, we optimize the LoRA parameters and masks with $\lambda_M{=}5\times10^{-5}$ and prune the masks every 0.1k iterations. We divide the denoising trajectory into four temporal stages based on the elbow-point analysis in Appendix~\ref{sec:appendix-number-of-stages} and apply the cached-feature teacher–student distillation described in Appendix~\ref{sec:appendix-teacher-student-distillation}. Per-pipeline hyperparameters and the shared calibration and training costs of SVDQuant, QAT, and TASQ are detailed in Appendix~\ref{sec:appendix-training-details}.

\paragraph{Datasets and Metrics.}
Training prompts are sampled from COCO Captions~\citep{chen2015microsoftcococaptionsdata}. For evaluation, we use 5K prompts each from MJHQ-30K~\citep{li2024playgroundv25} and sDCI~\citep{urbanek2024picture}, covering stylized and densely captioned image distributions. We report FID ($\downarrow$) and ImageReward ($\uparrow$) for quality, and LPIPS ($\downarrow$) and PSNR ($\uparrow$) for fidelity to the full-precision model, on both datasets.

% \subsection{Results}

% The gains appear on both DiT and U-Net backbones. For SDXL-Turbo, the four learned stages correspond exactly to its four denoising steps.

% \paragraph{Analysis of Temporal-Adaptiveness.}
% The learned precision follows the sensitivity profile in Figure~\ref{fig:avg}.The gains appear on both DiT and U-Net backbones. the noisiest stage generally receives more bits, later stages receive fewer, and the range within each stage shows that layers do not share the same requirement. A static allocation must follow the upper envelope of these demands; TASQ instead follows the demand at each stage and layer.
% The schedule-based sensitivity derived in Supplementary Section~D agrees closely with this allocation: its Pearson correlation with the learned bits is $0.97$ for PixArt-$\Sigma$ and $0.96$ for SDXL-Turbo (Supplementary Table~S4).
\paragraph{Analysis of Temporal-Adaptiveness.}
The learned precision follows the sensitivity profile in
Figure~\ref{fig:avg}. The noisiest stage generally receives more bits,
whereas later stages require fewer, and the variation within each stage
shows that layers do not share the same precision requirement. A static
allocation must follow the upper envelope of these demands, whereas TASQ
adapts the precision to each stage and layer.

\begin{figure}[t]
    \centering
    \includegraphics[width=\linewidth]{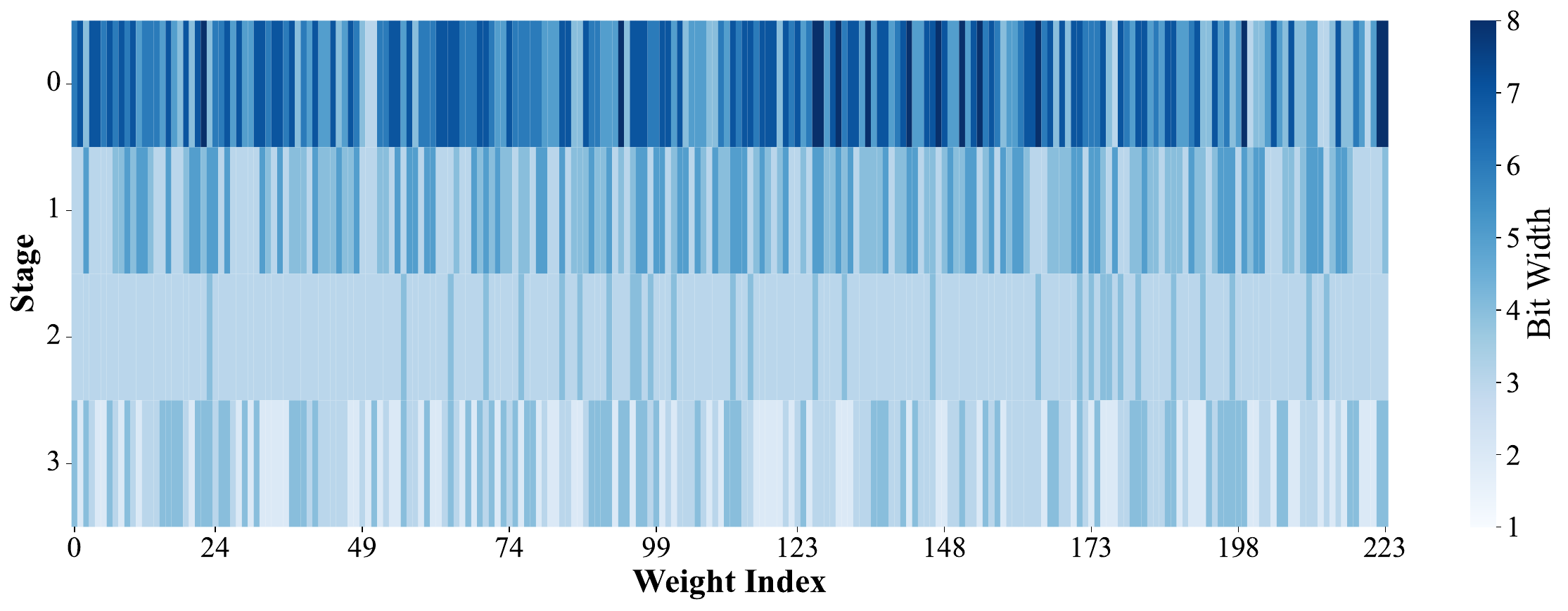}
    \vspace{-20pt}
    \caption{TASQ-learned operating precision across temporal stages and
    layers for PixArt-$\Sigma$ under W4A4.}
    \label{fig:pixart_heatmap}
\end{figure}

% The schedule-based sensitivity derived in Supplementary Section~D
% closely matches the learned allocation, achieving Pearson correlations
% of $0.97$ for PixArt-$\Sigma$ and $0.96$ for SDXL-Turbo
% (Supplementary Table~S4). The same trend is observed across both DiT
% and U-Net backbones.

% This trend follows the way quantization error propagates through the sampler. An error introduced near the noisy end affects more subsequent denoising updates and is amplified by the noise schedule before the final sample is produced. Later errors pass through fewer updates and can be represented with fewer weight bits. The result is not a single global decay, however. Attention and FFN layers peak at different parts of the trajectory, as Figure~\ref{fig:intro_sensitivity} shows, and the min--max bands in Figure~\ref{fig:avg} remain wide within each stage. TASQ therefore learns the broad temporal trend and the layer-specific departures from it. A timestep-only schedule would miss the latter, while a spatial-only schedule would average away the former.

This trend reflects how quantization error propagates through the denoising process. Errors introduced at early, noisy stages affect more subsequent updates and are further amplified by the noise schedule, whereas errors introduced later pass through fewer updates and can tolerate lower weight precision. However, this temporal trend is not uniform across layers. Attention and FFN layers are most sensitive at different parts of the trajectory, as shown in Figure~\ref{fig:intro_sensitivity}, while the wide min–max bands in Figure~\ref{fig:avg} indicate substantial layer-wise variation within each stage. TASQ therefore learns both the overall temporal trend and the layer-specific deviations. A timestep-only schedule cannot capture the latter, while a spatial-only schedule obscures the former.

\subsection{Ablation Study}

\paragraph{Effectiveness of Temporal–Spatial Adaptation.}
\begin{table*}[t]
\setlength{\tabcolsep}{6pt}
\caption{Ablation of temporal and spatial precision adaptation on PixArt-$\Sigma$. All settings store the same 8-bit model and use the same average 4-bit operating precision. \textbf{Temporal} and \textbf{Spatial} indicate whether precision varies across denoising stages and layers, respectively.}
\vspace{-9pt}
\label{tab:adaptivity}
\centering
\resizebox{\textwidth}{!}{%
\begin{tabular}{lccccccccccc}
\toprule
& & \multicolumn{2}{c}{Adaptation} & \multicolumn{4}{c}{MJHQ} & \multicolumn{4}{c}{sDCI} \\
\cmidrule(lr){3-4} \cmidrule(lr){5-8} \cmidrule(lr){9-12}
Method & \makecell{ Eff. W-Bit } & Temporal & Spatial
& FID ($\downarrow$) & IR ($\uparrow$) & LPIPS ($\downarrow$) & PSNR ($\uparrow$)
& FID ($\downarrow$) & IR ($\uparrow$) & LPIPS ($\downarrow$) & PSNR ($\uparrow$) \\
\midrule
Spatial only  & 4.00 & \xmark & \cmark
& 17.1 & 0.904 & 0.297 & 18.3
& 24.2 & 0.926 & 0.283 & 17.2 \\
Temporal only & 4.00 & \cmark & \xmark
& 18.4 & 0.909 & 0.290 & 18.5
& 25.3 & 0.929 & 0.333 & 17.0 \\
\rowcolor{irislight}
TASQ & 3.98 & \cmark & \cmark
& \textbf{16.1} & \textbf{0.932} & \textbf{0.282} & \textbf{18.6}
& \textbf{23.1} & \textbf{0.966} & \textbf{0.263} & \textbf{18.8} \\
\bottomrule
\end{tabular}}
\end{table*}

We examine whether temporal and spatial adaptation provide complementary benefits by comparing spatial-only, temporal-only, and joint temporal–spatial precision allocation. All settings store the same 8-bit model and use the same average 4-bit operating precision on PixArt-$\Sigma$. Spatial-only allocation achieves an FID of 17.1, while temporal-only allocation achieves 18.4. Combining both forms of adaptation improves the FID to \textbf{16.1}. These results show that allocating precision jointly across layers and denoising stages is more effective than adapting along either dimension alone.

\paragraph{Effect of the LSB Mask Design.}
Replacing the LSB mask with a sigmoid timestep gate increases the PixArt-$\Sigma$ W4A4 FID from 16.1 to 21.4 (Appendix~\ref{sec:appendix-lsb-mask-robustness}), suggesting that gradually suppressing the LSB signal before the model adapts is harmful.
% We likewise found that visiting stages sequentially
% caused earlier loss stagnation, motivating the farthest-stage-first
% training order in Section~\ref{sec:method-progressive}.

\paragraph{TASQ without SVDQuant Initialization.}
To determine whether the gain depends on the SVDQuant starting point,
Table~\ref{tab:no_svdquant} compares standard LoRA-QAT with TASQ when
both are trained directly from the same initialization. TASQ improves
ImageReward in every setting and lowers FID in all but one case,
indicating that the benefit comes from temporal adaptivity rather than
the initialization itself.

\begin{table}[t]
\centering
\scriptsize
\setlength{\tabcolsep}{2.6pt}
\vspace{-10pt}
\caption{TASQ versus non-temporal QAT without SVDQuant initialization.}
\vspace{-9pt}
\label{tab:no_svdquant}
\resizebox{\columnwidth}{!}{%
\begin{tabular}{llccccc}
\toprule
& & & \multicolumn{2}{c}{MJHQ} & \multicolumn{2}{c}{sDCI} \\
\cmidrule(lr){4-5}\cmidrule(lr){6-7}
Backbone & Setting & Method & FID ($\downarrow$) & IR ($\uparrow$) & FID ($\downarrow$) & IR ($\uparrow$) \\
\midrule
\multirow{4}{*}{PixArt-$\Sigma$}
& W4A8 & QAT & 18.5 & 0.908 & 25.0 & 0.940 \\
& W4A8 & \cellcolor{irislight}TASQ & \cellcolor{irislight}\textbf{17.2} & \cellcolor{irislight}\textbf{0.925} & \cellcolor{irislight}\textbf{24.3} & \cellcolor{irislight}\textbf{0.952} \\
& W4A4 & QAT & 19.0 & 0.885 & 25.5 & 0.922 \\
& W4A4 & \cellcolor{irislight}TASQ & \cellcolor{irislight}\textbf{17.5} & \cellcolor{irislight}\textbf{0.908} & \cellcolor{irislight}\textbf{24.6} & \cellcolor{irislight}\textbf{0.935} \\
\midrule
\multirow{4}{*}{SANA-1.6B}
& W4A8 & QAT & 17.5 & 1.025 & 22.8 & 1.010 \\
& W4A8 & \cellcolor{irislight}TASQ & \cellcolor{irislight}\textbf{17.0} & \cellcolor{irislight}\textbf{1.042} & \cellcolor{irislight}\textbf{22.5} & \cellcolor{irislight}\textbf{1.028} \\
& W4A4 & QAT & 19.2 & 0.940 & 27.0 & 0.855 \\
& W4A4 & \cellcolor{irislight}TASQ & \cellcolor{irislight}\textbf{18.2} & \cellcolor{irislight}\textbf{0.968} & \cellcolor{irislight}\textbf{26.0} & \cellcolor{irislight}\textbf{0.885} \\
\midrule
\multirow{4}{*}{SDXL-Turbo}
& W4A8 & QAT & 24.9 & 0.830 & \textbf{24.7} & 0.679 \\
& W4A8 & \cellcolor{irislight}TASQ & \cellcolor{irislight}\textbf{24.3} & \cellcolor{irislight}\textbf{0.838} & \cellcolor{irislight}25.0 & \cellcolor{irislight}\textbf{0.695} \\
& W4A4 & QAT & 25.1 & 0.810 & 25.4 & 0.665 \\
& W4A4 & \cellcolor{irislight}TASQ & \cellcolor{irislight}\textbf{24.0} & \cellcolor{irislight}\textbf{0.828} & \cellcolor{irislight}\textbf{24.9} & \cellcolor{irislight}\textbf{0.682} \\
\bottomrule
\end{tabular}}
\end{table}

\subsection{Efficiency Discussion}\label{sec:efficiency}
\paragraph{Operating regime.}
TASQ does not directly target reducing the stored model size.
The most sensitive stage still
determines the shared weight precision, which is stored once throughout
inference. Instead, TASQ reduces computation by executing less sensitive
stages with fewer weight bits. Table~\ref{tab:quality} therefore reports
Rel.\ BitOPs, not storage. On precision-scalable datapaths,
lower BitOPs translate directly into fewer execution cycles, whereas fixed
INT4/INT8 kernels cannot execute below their native precision.

\paragraph{Deployment procedure.}
Efficient deployment must consider both memory usage and inference speed. We therefore first choose the storage precision according to the available memory budget and then use TASQ to reduce the repeated denoising computation. The model is quantized and stored once at the highest precision that fits within the memory budget. TASQ is then trained to select among the lower precisions supported by the target device. The stored weight buffer remains unchanged throughout inference, and TASQ only determines how many bit planes each layer reads at each denoising stage. For example, an 8-bit deployment stores a single 8-bit model, while selected layers can execute at 3-bit or 2-bit precision by reading fewer bit planes. This procedure fixes the memory footprint while allowing TASQ to improve inference speed through stage- and layer-specific allocation.

\paragraph{Training efficiency.}
The masks add little optimization cost. With the same 5.0k-iteration schedule, TASQ matches QAT in memory and wall-clock time. Starting from SVDQuant, the masks and LoRA adapters require another 1.4k iterations: 0.5 hours on SANA-1.6B and 2.1 hours on SDXL-Turbo (Table~\ref{tab:train_efficiency}).

\begin{table}[t]
\centering
\scriptsize
\setlength{\tabcolsep}{3pt}
\caption{Training cost at W4A8. ``+TASQ'' is the additional adaptation after SVDQuant.}
\vspace{-9pt}
% \vspace{-10pt}
\label{tab:train_efficiency}
\resizebox{\columnwidth}{!}{%
\begin{tabular}{llcccc}
\toprule
Model & Method & Iter. & Time (h) & Mem. (GB) & FID \\
\midrule
\multirow{4}{*}{SANA-1.6B}
& SVDQuant & -- & 6.8 & 18.6 & 17.7 \\
& \cellcolor{irislight}+TASQ & \cellcolor{irislight}+1.4k & \cellcolor{irislight}+0.5 & \cellcolor{irislight}9.8 & \cellcolor{irislight}\textbf{15.9} \\
& QAT & 5.0k & 1.8 & 9.1 & 17.5 \\
& \cellcolor{irislight}TASQ & \cellcolor{irislight}5.0k & \cellcolor{irislight}1.8 & \cellcolor{irislight}9.1 & \cellcolor{irislight}17.0 \\
\midrule
\multirow{4}{*}{SDXL-Turbo}
& SVDQuant & -- & 10.4 & 12.4 & 24.5 \\
& \cellcolor{irislight}+TASQ & \cellcolor{irislight}+1.4k & \cellcolor{irislight}+2.1 & \cellcolor{irislight}16.7 & \cellcolor{irislight}\textbf{23.8} \\
& QAT & 5.0k & 8.2 & 11.6 & 24.9 \\
& \cellcolor{irislight}TASQ & \cellcolor{irislight}5.0k & \cellcolor{irislight}8.2 & \cellcolor{irislight}11.6 & \cellcolor{irislight}24.3 \\
\bottomrule
\end{tabular}}
\end{table}

\begin{table}[t]
\centering
\small
\setlength{\tabcolsep}{4pt}
\vspace{-10pt}
\caption{Measured four-tile GEMM cost on the Temporal-Precision Engine with A4. The mixed schedule incurs no precision-switching overhead.}
\vspace{-9pt}
\label{tab:measured}
\resizebox{\linewidth}{!}{%
\begin{tabular}{lcccccc>{\columncolor{irislight}}c}
\toprule
& \multicolumn{6}{c}{Static (one precision $\times$ 4 tiles)} & \textbf{Mixed} \\
\cmidrule(lr){2-7}
4 tiles, A$=$4 & W1 & W2 & W3 & W4 & W6 & W8 & $\mathbf{1,2,3,4}$ \\
\midrule
Cycles & 528 & 1{,}056 & 1{,}584 & 2{,}112 & 3{,}168 & 4{,}224 & \textbf{1{,}320} ($+0$) \\
Latency ($\mu$s) & 2.89 & 5.78 & 8.67 & 11.57 & 17.35 & 23.13 & \textbf{7.23} \\
Energy ($\mu$J) & 1.89 & 3.78 & 5.67 & 7.56 & 11.35 & 15.13 & \textbf{4.73} \\
\bottomrule
\end{tabular}}
\end{table}

\begin{figure}[t]
    \centering
    \includegraphics[width=\linewidth]{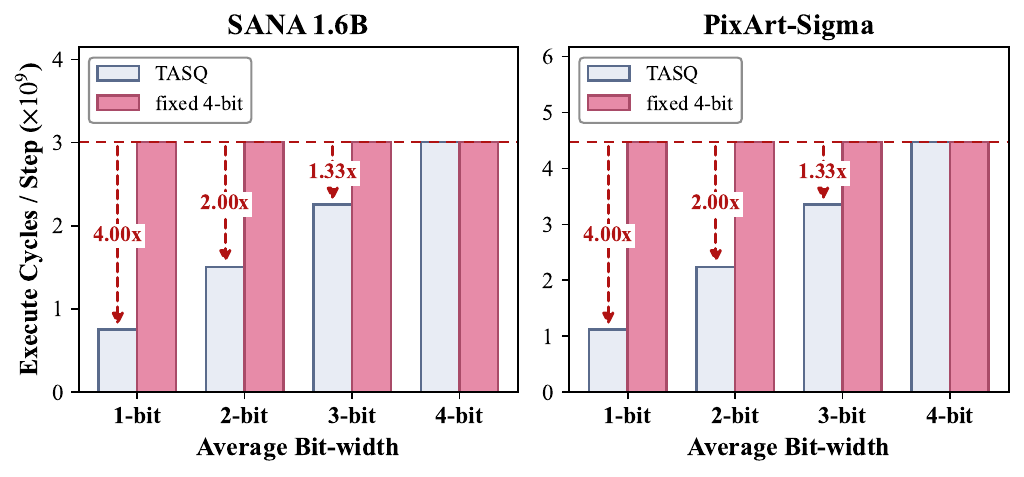}
     \vspace{-22pt}
    \caption{Execute cycles per denoising step under the measured bit-serial cycle law: TASQ executes at its average effective bit-width, while a fixed 4-bit datapath cannot descend below 4-bit.}
    \label{fig:cycles_bar}
\end{figure}

\paragraph{Inference efficiency.}
We measure bit-serial GEMM on our Temporal-Precision Engine, whose execution cost scales with the selected precision. As shown in Table~\ref{tab:measured}, execution cycles scale linearly with the product of weight and activation bit widths, while switching precision introduces no measured cycle overhead. The execution cost of a mixed-precision schedule therefore equals the sum of the measured cycles at each precision. Under this measured cost model, TASQ retains static W8 quality while reducing execution cycles by approximately $25\%$, corresponding to an average operating weight precision of about 6 bits. At the same average W4A4 computational budget, TASQ improves MJHQ FID by $1.0$--$3.1$ points over static quantization. The Temporal-Precision Engine therefore realizes TASQ's adaptive precision schedule with execution time proportional to the selected precision and no measured overhead for precision switching.

\section{Conclusion}
We introduced TASQ, a temporal weight-precision method that separates stored precision from the precision used during denoising. Its stage- and layer-specific LSB masks operate on one shared buffer, avoiding duplicated checkpoints while adapting computation to the denoising trajectory. Across DiT and U-Net backbones, TASQ retained the quality of static 8-bit quantization with less computation and improved generation quality under the same average compute budget. The Temporal-Precision Engine further translated the reduced bit-plane computation into measured cycle savings on precision-scalable hardware. These results show that temporal weight allocation improves the quality--compute trade-off while preserving a single shared weight representation.

{\small
\bibliography{tasq}}

% Start the appendix on a fresh page (\clearpage, not \newpage:
% in two-column mode \newpage only breaks the column).
\clearpage
\appendix
\section{Implementation Details}
\label{sec:appendix-implementation-details}

\subsection{Quantization}
\label{sec:appendix-quantization}

\subsubsection{Weight Quantization}
The quantization process is formally defined in Equation~\eqref{eq:weight_quant}:
\begin{equation}
Q(w,n) = \operatorname{clamp}\left(\left\lfloor \frac{w}{\alpha} + z \right\rfloor, 0, 2^n - 1\right),
\label{eq:weight_quant}
\end{equation}
where $\lfloor \cdot \rfloor$ denotes the floor operation. We use $2^n$, rather than $2^n{-}1$, in the scale so that the quantization grid aligns with the truncation boundaries:
\begin{equation}
\alpha = \frac{\max(w) - \min(w)}{2^n}, \qquad z = -\frac{\min(w)}{\alpha}.
\label{eq:weight_scale}
\end{equation}
The corresponding dequantization is
\begin{equation}
\hat{w} = \frac{2^n}{2^n - 1} \cdot \alpha \cdot (Q(w,n) - z).
\label{eq:weight_dequant}
\end{equation}

\subsubsection{Activation Quantization}
The quantization process is formally defined in Equation~\eqref{eq:act_quant}:
\begin{equation}
Q(x,n) = \operatorname{clamp}\left(\left\lfloor \frac{x}{\alpha} + z \right\rceil, 0, 2^n - 1\right),
\label{eq:act_quant}
\end{equation}
where $\lfloor \cdot \rceil$ denotes the rounding-to-nearest operation. We calculate the scaling factor $\alpha$ using the denominator $2^n - 1$ to fully utilize the representable range of the $n$-bit integer:
\begin{equation}
\alpha = \frac{\max(x) - \min(x)}{2^n - 1}, \qquad z = -\frac{\min(x)}{\alpha}.
\label{eq:act_scale}
\end{equation}
The activation is dequantized as
\begin{equation}
\hat{x} = \alpha \cdot (Q(x,n) - z).
\label{eq:act_dequant}
\end{equation}

\subsection{Teacher--Student Distillation}
\label{sec:appendix-teacher-student-distillation}
We sample training prompts from MS-COCO and precompute their conditional features $c_t$ and unconditional features $uc_t$. The student can then be trained without keeping the teacher, VAE, and text encoder active in memory.

\subsection{Overall Training Algorithm}
\label{ssec:appendix-training-algorithm}
The overall training algorithm is summarized in Algorithm~\ref{alg:tasq}.

\begin{algorithm}[t]
\caption{Overall Training Algorithm of TASQ}
\label{alg:tasq}
\begin{algorithmic}[1]
\Require training data $X$, total diffusion timesteps $T$, number of temporal stages $n$, pruning interval $I$
\Ensure quantized model $G$
\State Initialize parameters $F_1$, $F_2$, $m_{t,l}$
\State Initialize regularization strength $\lambda_M$
\State Divide $T$ into $n$ stages: $T = \{T_1, T_2, \ldots, T_n\}$
\State Initialize mask training step $s = 0$, mask training interval $s_0 = I/n$
\For{iteration $= 0 \ldots i$}
    \State Sample a timestep $t \sim \mathcal{U}(T)$ corresponding to current stage
    \State Compute $M_{t,l} = \mathcal{G}(s,m_{t,l})$
    \State LSB masked weight: $Q_{M,n}^{(t,l)} = Q_n - (1 - M_{t,l}) \cdot \mathrm{LSB}$
    \Statex \hspace{1.5em}Teacher--Student Distillation
    \State Quantized model $\epsilon_q = F_q(x_t,t;Q_{M,n}^{(t,l)})$
    \State Full-precision teacher $\epsilon_{\mathrm{fp}} = F_{\mathrm{fp}}(x_t,t;W)$
    \State Distillation loss: $L_{\mathrm{distill}} = \lVert \epsilon_q - \epsilon_{\mathrm{fp}} \rVert^2$
    \State Total loss: $L_{\mathrm{total}} = L_{\mathrm{distill}} + \lambda_M \sum_{t,l} M_{t,l}$
    \State Update LoRA weights $F_1$, $F_2$ and mask parameter $m_{t,l}$ via gradient descent
    \If{iteration $> 0$ and iteration \% $n == 0$}
        \State $s = s + 1$
    \EndIf
    \If{$s == s_0$}
        \If{$m_{t,l} < 0$}
            \State Prune corresponding LSB in layer $l$ at timestep $t$
            \State reset $m_{t,l} = 0$
        \EndIf
        \State reset $s = 0$
    \EndIf
\EndFor
\end{algorithmic}
\end{algorithm}

\section{Robustness of the LSB Mask}
\label{sec:appendix-lsb-mask-robustness}
Table~\ref{tab:mask_formulation} compares our Continuous LSB Mask (CLM) with the temperature-controlled sigmoid gate used in CSQ~\citep{csq}. A standard sigmoid gate starts at 0.5, so it attenuates every LSB at the beginning of training. CLM instead starts at one, keeps the LSBs intact during the initial updates, and later anneals each gate toward zero or one. On PixArt-$\Sigma$ W4A4, this change lowers FID from 21.4 to 16.1 on MJHQ and from 27.8 to 23.1 on sDCI.

\begin{table*}[t]
\caption{Mask formulation on PixArt-$\Sigma$. Initializing CLM at one avoids attenuating LSBs at the start of training.}
\label{tab:mask_formulation}
\scriptsize
\centering
\resizebox{\textwidth}{!}{%
\begin{tabular}{cccccccccccc}
\toprule
Backbone & Model & (Avg) Precision & Mask Type & FID ($\downarrow$) & IR ($\uparrow$) & LPIPS ($\downarrow$) & PSNR ($\uparrow$) & FID ($\downarrow$) & IR ($\uparrow$) & LPIPS ($\downarrow$) & PSNR ($\uparrow$) \\
\midrule
\multirow{2}{*}{DiT} & \multirow{2}{*}{\makecell{PixArt-$\Sigma$\\(20 Steps)}} & INT W4A4 & Sigmoid & 21.4 & 0.863 & 0.372 & 15.9 & 27.8 & 0.882 & 0.398 & 14.6 \\
& & INT W4A4 & CLM (Ours) & \textbf{16.1} & \textbf{0.932} & \textbf{0.282} & \textbf{18.6} & \textbf{23.1} & \textbf{0.966} & \textbf{0.263} & \textbf{18.8} \\
\bottomrule
\end{tabular}}
\end{table*}

\section{Training Details}
\label{sec:appendix-training-details}

All operating points share the same $b_{\max}{=}8$ buffer design and are compared at matched average effective bit-width (BitOPs), each fine-tuned for its setting; under a memory constraint, $b_{\max}$ is instead set to the affordable precision, giving storage identical to a native model at that width while TASQ still executes below it on average.

This section details the training configuration used for each pipeline evaluated
above. We separate the description into three parts: (i) the optimization
hyperparameters of the fine-tuning loop (Table~\ref{tab:training_setting}),
(ii) the calibration / training data budget per backbone
(Table~\ref{tab:calibration_dataset}), and (iii) the resulting training cost in
wall-clock time and peak memory (Table~\ref{tab:train_efficiency}).

\paragraph{Optimization hyperparameters.}
Table~\ref{tab:training_setting} compares the four pipelines studied above
along four axes: total iterations, LSB pruning interval (applicable only to TASQ-based
pipelines), residual LoRA rank, and batch size. Pure QAT and pure TASQ are trained
from the FP-initialized model for 5.0k iterations, whereas SVDQuant-initialized
variants ({SVDQuant+QAT}, {SVDQuant+TASQ}) start from a calibrated SVDQuant checkpoint
and only require 1.4k additional fine-tuning iterations. TASQ-based pipelines
additionally maintain the temporal--spatial LSB mask, which is annealed and pruned
every 0.1k iterations. The trainable residual LoRA branch is fixed at rank 32 across
all variants; SVDQuant-initialized pipelines additionally carry a frozen SVD low-rank
branch (rank 32 for the 4-bit setting, rank 16 for the 6-bit setting) inherited from
the PTQ initialization.

\begin{table}[t]
\centering
\small
\caption{Training-setting comparison across QAT, TASQ, SVDQuant+QAT, and SVDQuant+TASQ.
``Pruning Int.'' is the LSB pruning interval, applicable only to TASQ-based pipelines.
``Residual LoRA Rank'' refers to the trainable LoRA branch; SVDQuant-initialized
pipelines additionally carry a frozen SVD low-rank branch (rank 32 for 4-bit, rank 16
for 6-bit) inherited from the PTQ initialization.}
\label{tab:training_setting}
\setlength{\tabcolsep}{8pt}
\resizebox{\columnwidth}{!}{%
\begin{tabular}{lcccc}
\toprule
Method            & Iter. & Pruning Int. & Trainable LoRA Rank & Batch Size \\
\midrule
QAT               & 5.0k  & --           & 32                 & 4 \\
TASQ              & 5.0k  & 0.1k         & 32                 & 4 \\
SVDQuant + QAT    & 1.4k  & --           & 32                 & 4 \\
SVDQuant + TASQ   & 1.4k  & 0.1k         & 32                 & 4 \\
\bottomrule
\end{tabular}}
\end{table}

\paragraph{Calibration / training data.}
To ensure a controlled comparison, all three methods (SVDQuant, QAT, TASQ) draw from
the same pool of $128 \times T$ (prompt, timestep) supervision pairs, where $T$ is the
number of denoising steps used by each backbone. Concretely, we sample 128 distinct
text prompts from MS-COCO Captions and pre-compute the conditional and unconditional
teacher features $c_t$ and $\mathrm{uc}_t$ at every denoising timestep, yielding the
cached pool described in Appendix~\ref{sec:appendix-teacher-student-distillation}.
SVDQuant uses this pool for scale calibration and the SVD decomposition of its
low-rank branch; QAT and TASQ sample from it for cached teacher--student distillation.
Table~\ref{tab:calibration_dataset} summarizes the pool size for each backbone.

\begin{table}[t]
\centering
\small
\caption{Calibration / training data budget for SVDQuant, QAT, and TASQ on each
backbone. The pool size is $128 \times T$, where $T$ is the number of denoising steps.
All three methods share the identical pool to isolate the contribution of the
optimization procedure from any difference in supervision volume.}
\label{tab:calibration_dataset}
\setlength{\tabcolsep}{8pt}
\resizebox{\columnwidth}{!}{%
\begin{tabular}{lcccc}
\toprule
Model            & Steps ($T$) & SVDQuant                      & QAT                           & TASQ                          \\
\midrule
PixArt-$\Sigma$  & 20          & $128 \times 20 = 2{,}560$     & $128 \times 20 = 2{,}560$     & $128 \times 20 = 2{,}560$     \\
SANA-1.6B        & 20          & $128 \times 20 = 2{,}560$     & $128 \times 20 = 2{,}560$     & $128 \times 20 = 2{,}560$     \\
SDXL-Turbo       & 4           & $128 \times 4 = 512$          & $128 \times 4 = 512$          & $128 \times 4 = 512$          \\
\bottomrule
\end{tabular}}
\end{table}

\paragraph{Training cost.}
Table~\ref{tab:train_efficiency} reports wall-clock time, peak memory, and the
resulting FID for every pipeline at W4A8. Pure TASQ matches pure QAT exactly in
both time and memory at the same 5.0k iterations, so the temporal--spatial masks
add no measurable optimization cost over a non-temporal LoRA-QAT baseline. Starting
from a calibrated SVDQuant checkpoint, TASQ needs 1.4k further iterations, which
costs $+0.5$ hours on SANA-1.6B and $+2.1$ hours on SDXL-Turbo and improves FID
from $17.7$ to $15.9$ and from $24.5$ to $23.8$ respectively. Peak memory during
the SVDQuant-initialized adaptation is lower than SVDQuant's own calibration pass
on SANA-1.6B ($9.8$ against $18.6$~GB) because only the LoRA branch and the masks
carry gradients.

\section{Interpretation of Timestep Sensitivity}
\label{sec:appendix-timestep-sensitivity-interpretation}

\subsection{Global Timestep Sensitivity}
\label{ssec:appendix-global-sensitivity}

This section provides a mechanism-level interpretation of the timestep sensitivity observed in Figure~\ref{fig:avg}. Following the denoising order used by TASQ, Stage~0 denotes the noisiest timesteps (large $t$), whereas the final stage corresponds to the cleanest timesteps. Under this convention, the theoretically derived sensitivity aligns almost perfectly with TASQ's learned bit allocation.

\paragraph{Error propagation in quantized DDIM.}
The DDIM update can be written as
\begin{equation}
\resizebox{\columnwidth}{!}{$\displaystyle
 x_{t-1} = \frac{\sqrt{\bar{\alpha}_{t-1}}}{\sqrt{\bar{\alpha}_{t}}} x_{t} + B_{t} \, \epsilon_{\theta}(x_{t}, t),
 \quad
 B_{t} = \sqrt{1-\bar{\alpha}_{t-1}} - \frac{\sqrt{\bar{\alpha}_{t-1}(1-\bar{\alpha}_{t})}}{\sqrt{\bar{\alpha}_{t}}}.
$}
\label{eq:appendix_ddim_update}
\end{equation}
When the denoiser is quantized as $\tilde{\epsilon}_{\theta} = \epsilon_{\theta} + \vartheta_{t}$, defining the cumulative deviation $\delta_{t} := \tilde{x}_{t} - x_{t}$ with $\delta_{T} = 0$, a first-order Taylor expansion yields the recursive error propagation
\begin{equation}
 \delta_{t-1} = A_{t} \, \delta_{t} + B_{t} \, \vartheta_{t},
 \qquad
 A_{t} = \frac{\sqrt{\bar{\alpha}_{t-1}}}{\sqrt{\bar{\alpha}_{t}}} I + B_{t} J_{x_{t}},
\label{eq:appendix_error_recursion}
\end{equation}
where $J_{x_t}$ denotes the Jacobian of $\epsilon_{\theta}(x_t,t)$ with respect to $x_t$.

\paragraph{Closed-form final output error.}
Unrolling Equation~\eqref{eq:appendix_error_recursion} from $\delta_{T}=0$ to $\delta_{0}$, following the closed-form DDIM error propagation analysis of \cite{liu2025error}, gives
\begin{equation}
 \delta_{0} = \sum_{k=1}^{T} \underbrace{\left(\prod_{j=1}^{k-1} A_{j}\right) B_{k}}_{C_{k}} \, \vartheta_{k}.
\label{eq:appendix_closed_form_error}
\end{equation}
Each coefficient $C_k$ captures how the per-step quantization error $\vartheta_k$ injected at timestep $k$ is amplified through the subsequent denoising trajectory before reaching the final output $x_0$.

\paragraph{Closed-form timestep sensitivity.}
Applying the approximation $J_{x_t} \approx 0$ used in prior analyses of quantized diffusion error propagation~\citep{liu2025error}, each $A_t$ reduces to $\frac{\sqrt{\bar{\alpha}_{t-1}}}{\sqrt{\bar{\alpha}_{t}}} I$, and the product telescopes as
\begin{equation}
 \prod_{j=1}^{k-1} A_{j}
 \approx
 \frac{\sqrt{\bar{\alpha}_{0}}}{\sqrt{\bar{\alpha}_{k-1}}} I
 \approx
 \frac{1}{\sqrt{\bar{\alpha}_{k-1}}} I.
\label{eq:appendix_telescope}
\end{equation}
This yields the following closed-form timestep sensitivity that depends only on the noise schedule:
\begin{equation}
 \mathcal{S}(k) := \lVert C_{k} \rVert \approx \frac{|B_{k}|}{\sqrt{\bar{\alpha}_{k-1}}}.
\label{eq:appendix_timestep_sensitivity}
\end{equation}
The noise schedule makes both factors larger near the noisy end: $\bar{\alpha}_{k-1}$ becomes smaller, while $|B_k|$ increases. An error introduced there is therefore amplified more strongly and propagates through more subsequent steps before reaching $x_0$. Errors introduced near the clean end have fewer steps in which to accumulate.

\paragraph{Verification against TASQ bit allocation.}
We evaluate $\mathcal{S}(k)$ on each model's noise schedule and average it within the same stages used by TASQ. Table~\ref{tab:timestep_sensitivity} reports Pearson correlations of $0.97$ for PixArt-$\Sigma$ and $0.96$ for SDXL-Turbo between sensitivity and learned precision. Stage~0 is the most sensitive stage in both models and receives the most bits.

\begin{table}[t]
\centering
\scriptsize
\setlength{\tabcolsep}{2pt}
\caption{Stage-wise noise-schedule sensitivity and learned weight precision. Stage~0 contains the noisiest timesteps.}
\label{tab:timestep_sensitivity}
\resizebox{\columnwidth}{!}{%
\begin{tabular}{llcc}
\toprule
Model & Stage & Sensitivity & TASQ bits \\
\midrule
\multirow{4}{*}{PixArt-$\Sigma$}
& 0 (noisy) & \textbf{0.71649} & \textbf{5.87} \\
& 1 & 0.06710 & 3.99 \\
& 2 & 0.01131 & 3.13 \\
& 3 (clean) & 0.00396 & 3.01 \\
\midrule
\multirow{4}{*}{SDXL-Turbo}
& 0 (noisy) & \textbf{0.08837} & \textbf{4.53} \\
& 1 & 0.01734 & 4.05 \\
& 2 & 0.00537 & 3.71 \\
& 3 (clean) & 0.00259 & 3.71 \\
\bottomrule
\end{tabular}}
\end{table}

\subsection{Layer-wise Timestep Sensitivity}
\label{ssec:appendix-layerwise-sensitivity}
\begin{figure*}[t]
    \centering
    \includegraphics[width=\textwidth]{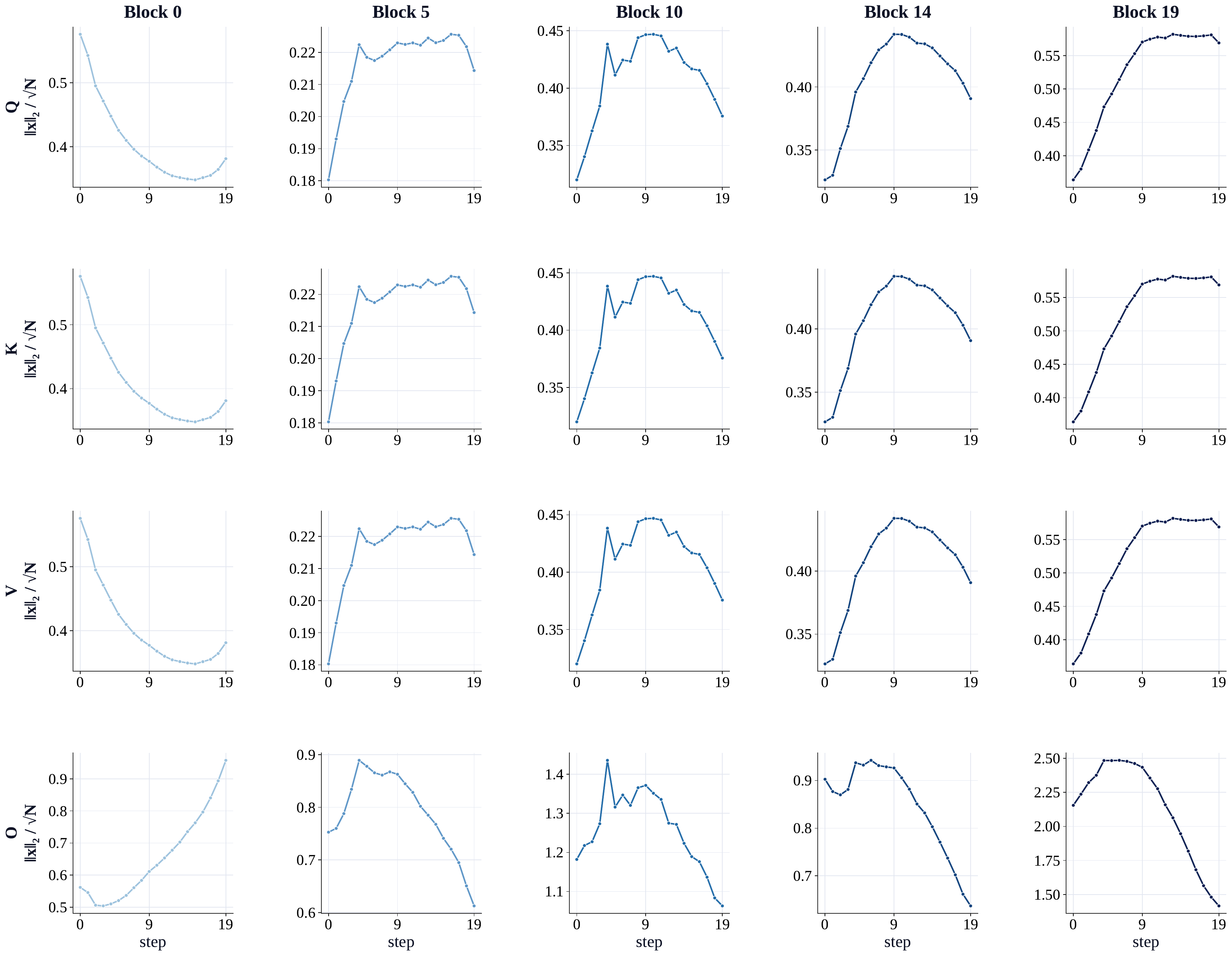}
    \caption{
        Per-projection $\times$ per-block input-activation RMS,
        $\lVert x\rVert_2/\sqrt{N}$, of SANA-1.6B across denoising steps.
    }
    \label{fig:layer_timestep_sensitivity}
\end{figure*}

\noindent
Figure~\ref{fig:layer_timestep_sensitivity} visualizes the activation dynamics of
SANA-1.6B over 20 denoising steps, averaged over 10 prompts. Rows correspond to
the four self-attention projections, Q, K, V, and O, while columns show five
evenly spaced transformer blocks. Q, K, and V share the residual-branch input
and therefore exhibit nearly identical trajectories. In contrast, the output
projection O receives the post-attention hidden state and shows both larger
activation magnitude and stronger step-to-step variation.

The activation trajectory also changes substantially with depth. In early
blocks, the RMS tends to decay monotonically across denoising, whereas deeper
blocks can rise sharply on the noisy side and peak around intermediate
timesteps. This depth- and projection-conditioned non-stationarity motivates
per-stage quantization, since a single static scale may fail to cover both ends
of the activation distribution, especially in deeper output projections.

\subsection{Learned Bit Allocation across Layers and Stages}
\label{ssec:appendix-bit-allocation-maps}

\begin{figure*}[p]
    \centering
    \begin{subfigure}{\textwidth}
        \centering
        \includegraphics[width=0.86\textwidth]{figures/supp/pixart-4bit_heatmap.pdf}
        \caption{PixArt-$\Sigma$ (0.6B, DiT, 20 steps)}
        \label{fig:bitmap_pixart}
    \end{subfigure}
    \\[6pt]
    \begin{subfigure}{\textwidth}
        \centering
        \includegraphics[width=0.86\textwidth]{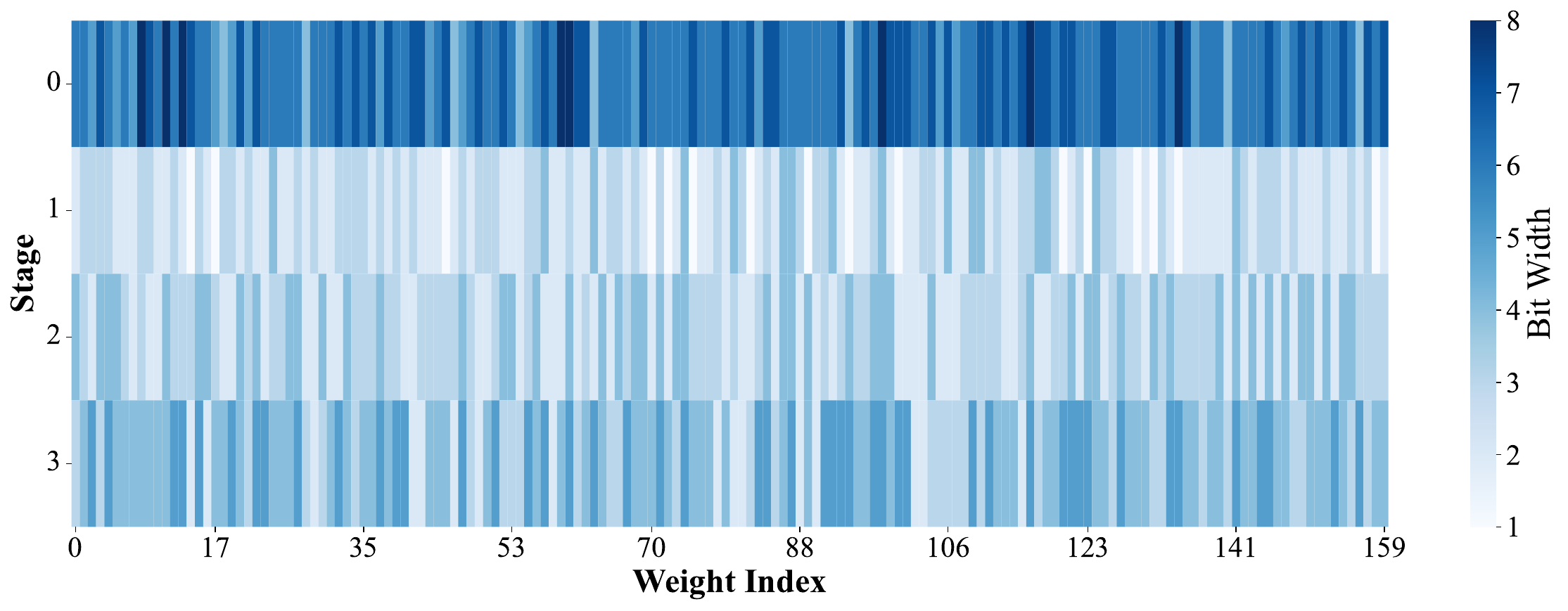}
        \caption{SANA-1.6B (DiT, 20 steps)}
        \label{fig:bitmap_sana}
    \end{subfigure}
    \\[6pt]
    \begin{subfigure}{\textwidth}
        \centering
        \includegraphics[width=0.86\textwidth]{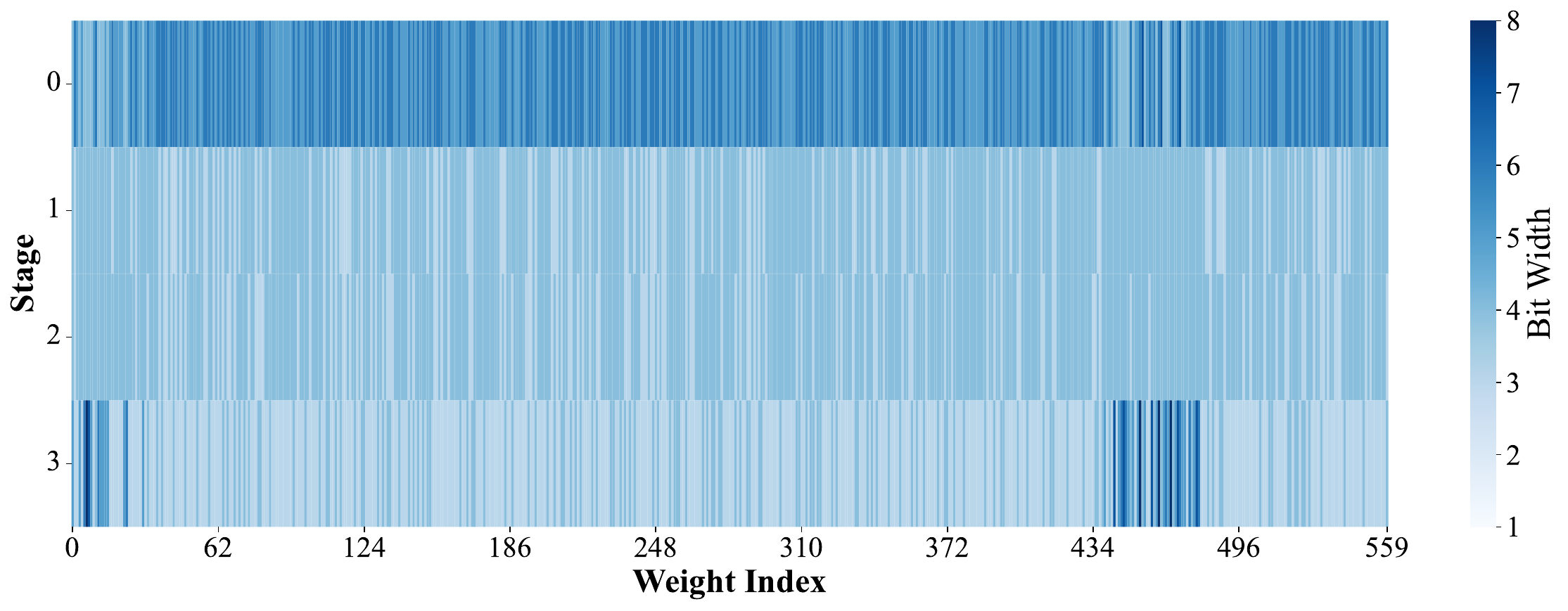}
        \caption{SDXL-Turbo (2.6B, U-Net, 4 steps)}
        \label{fig:bitmap_sdxl}
    \end{subfigure}
    \caption{TASQ-learned operating weight precision across temporal stages
    (rows) and weight tensors (columns) at the average W4A4 setting. Darker is
    higher precision. Stage~0 is the noisiest stage. All three backbones keep
    the most planes at Stage~0, but the allocation within each stage varies
    strongly across tensors, and the per-stage ordering is not monotonic in the
    denoising direction.}
    \label{fig:appendix_bit_allocation}
\end{figure*}

Figure~\ref{fig:appendix_bit_allocation} shows the operating precision TASQ
learns for every quantized weight tensor at every temporal stage, for all three
backbones at the average W4A4 setting. The main paper reproduces the
PixArt-$\Sigma$ panel; the SANA-1.6B and SDXL-Turbo panels are given here.

Three properties are visible in all three models. First, Stage~0, the noisiest
stage, is uniformly the darkest row: it retains the most least-significant
planes, matching the closed-form schedule sensitivity of
\S\ref{ssec:appendix-global-sensitivity}. Second, precision does not decay
monotonically along the trajectory. In PixArt-$\Sigma$ the lightest row is
Stage~2 rather than Stage~3, and in SDXL-Turbo Stages~1 and~2 are close to each
other, so the stage-wise averages in Table~\ref{tab:timestep_sensitivity} hide
non-monotonic structure that a hand-designed decay schedule would miss. Third,
within every row the allocation varies strongly from tensor to tensor, and the
high-precision tensors are not the same ones across stages. This is the
component that a purely temporal schedule cannot express: it would have to
assign one precision to an entire row, and therefore follow the upper envelope
of that row rather than its actual distribution.

The number of weight indices differs per backbone ($224$ for PixArt-$\Sigma$,
$160$ for SANA-1.6B, $560$ for SDXL-Turbo) because it counts the quantized
linear and convolutional tensors of each architecture. Indices are ordered by
network depth.

\section{Additional Comparison with Temporal-Awareness}
\label{sec:appendix-temporal-awareness-comparison}

Table~\ref{tab:adaptivity} established that temporal and spatial adaptation are complementary, and Table~\ref{tab:no_svdquant} that the gain survives without SVDQuant initialization. This section adds the two comparisons that do not fit there: Table~\ref{tab:positioning} places TASQ against timestep-aware quantizers along the axes that distinguish them, and Table~\ref{tab:cifar10_quant} compares against timestep-aware baselines on CIFAR-10.

\begin{table}[t]
\centering
\scriptsize
\setlength{\tabcolsep}{3pt}
\renewcommand{\arraystretch}{1.05}
\caption{Positioning against timestep-aware quantization: only TASQ adapts \emph{weight} precision per denoising step (W), learns the allocation end-to-end, keeps a single shared weight buffer ($1{\times}$ store), and measures the resulting compute saving on a bit-serial accelerator. \footnotesize$^\S$single per-layer allocation; $^\dagger$per-layer $+$ temporal distillation; $^\ddagger$activation \emph{ranges}, not precision; ``--'' marks methods whose allocation is fixed across denoising steps, for which the per-step storage question does not arise.}
\label{tab:positioning}
\resizebox{\columnwidth}{!}{%
\begin{tabular}{lccccc}
\toprule
& \multicolumn{2}{c}{Per-step prec.} & & & \\
\cmidrule(lr){2-3}
Method & W & A & Learned & $1{\times}$ store & HW meas. \\
\midrule
MPQ-Diff$^\S$~\shortcite{MPQdiff}      & \xmark & \xmark & \xmark & -- & \xmark \\
MPQ-DM$^\dagger$~\shortcite{MPQdm}     & \xmark & \xmark & \xmark & -- & \xmark \\
TCAQ-DM$^\ddagger$~\shortcite{TCAQdm}  & \xmark & \xmark & \xmark & \cmark & \xmark \\
AdaTSQ~\shortcite{AdaTSQ}             & \xmark & \cmark & \xmark & \cmark & \xmark \\
\rowcolor{irislight}\textbf{TASQ (ours)} & \textbf{\cmark} & \xmark & \cmark & \cmark & \cmark \\
\bottomrule
\end{tabular}}
\end{table}

\begin{table}[t]
\centering
\setlength{\tabcolsep}{3pt}
\scriptsize
\caption{CIFAR-10 results for a 100-step DDIM model. TASQ matches the best IS and gives the lowest FID at both activation precisions.}
\label{tab:cifar10_quant}
\resizebox{\columnwidth}{!}{%
\begin{tabular}{lcccc}
\toprule
Method & A-bit & W-bit & IS ($\uparrow$) & FID ($\downarrow$) \\
\midrule
Full Precision & 32 & 32 & 9.12 & 4.14 \\
\midrule
PTQ4DM & \multirow{5}{*}{8} & 4.00 & 9.31 & 10.12 \\
Q-Diffusion & & 4.00 & 9.12 & 4.93 \\
TFMQ-DM & & 4.00 & 9.13 & 4.78 \\
EfficientDM & & 4.00 & \textbf{9.41} & 3.80 \\
\rowcolor{irislight}\textbf{TASQ} & & 3.91 & \textbf{9.41} & \textbf{3.65} \\
\midrule
PTQ4DM & \multirow{5}{*}{4} & 4.00 & 0.45 & 375.12 \\
Q-Diffusion & & 4.00 & 0.71 & 384.21 \\
TFMQ-DM & & 4.00 & 3.19 & 236.63 \\
EfficientDM & & 4.00 & 9.37 & 3.91 \\
\rowcolor{irislight}\textbf{TASQ} & & 3.92 & \textbf{9.40} & \textbf{3.90} \\
\bottomrule
\end{tabular}}
\end{table}
Table~\ref{tab:cifar10_quant} compares TASQ, implemented on EfficientDM, with PTQ4DM, Q-Diffusion~\citep{li2023qdiffusion}, TFMQ-DM~\citep{huang2024tfmqdm}, and EfficientDM on a 100-step CIFAR-10 DDIM model. The baselines keep weight precision fixed across timesteps, whereas TASQ learns a timestep-dependent weight allocation. At A8, TASQ matches EfficientDM's IS of 9.41 and lowers FID from 3.80 to 3.65. At A4, it raises IS from 9.37 to 9.40 and lowers FID from 3.91 to 3.90.

\section{Number of Stages}
\label{sec:appendix-number-of-stages}

\begin{figure*}[t]
    \centering
    \includegraphics[width=0.85\textwidth]{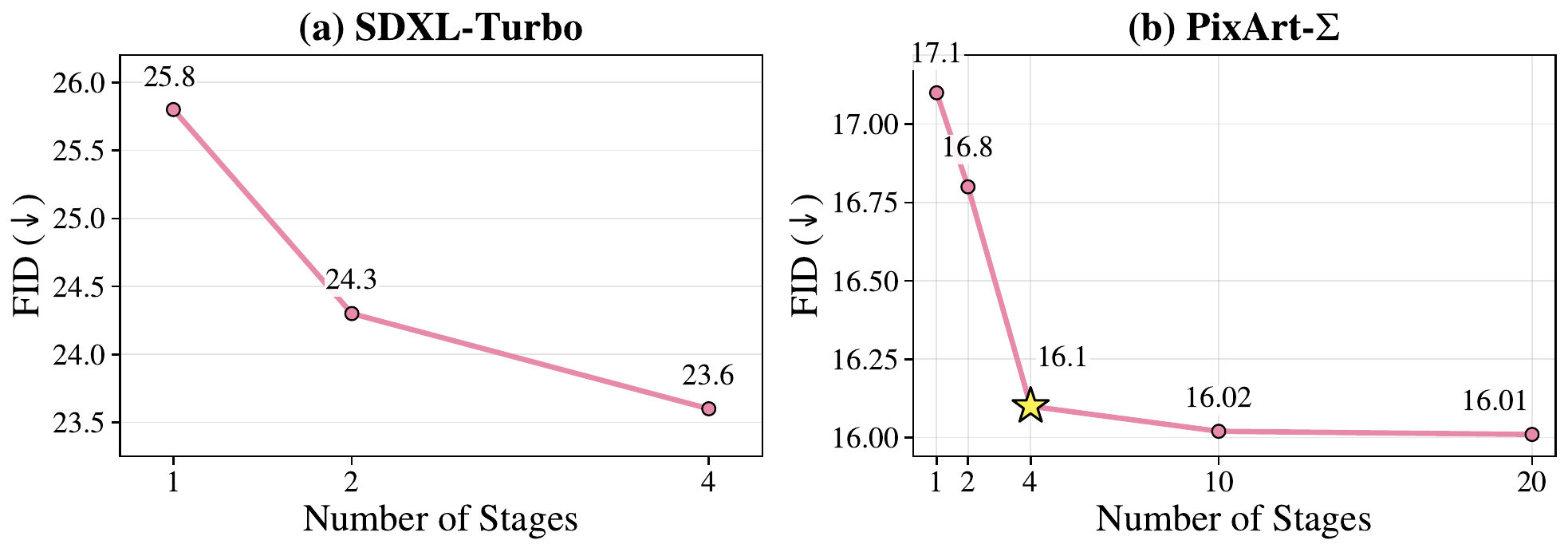}
    \caption{Ablation on the number of temporal stages on MJHQ for (a) SDXL-Turbo and (b) PixArt-$\Sigma$. For SDXL-Turbo, performance improves steadily up to four stages, matching its four denoising steps. For PixArt-$\Sigma$, four stages form the elbow point, after which the gains become marginal.}
    \label{fig:appendix_num_stages}
\end{figure*}

Figure~\ref{fig:appendix_num_stages} varies the number of stages on MJHQ. SDXL-Turbo improves as the stage count increases from one to four, matching its four denoising steps. PixArt-$\Sigma$ also improves up to four stages, with little change beyond that point. We therefore use four stages in the remaining experiments.

\section{Hardware Deployment Details}
\label{sec:appendix-hardware-deployment-details}

This section gives the deployment details behind the bit-serial measurements in
\S\ref{sec:efficiency}:
we give the bit-plane schedules the engine executes, describe how the operating
precision maps to the native precision set of a target device, and discuss the
restriction imposed by static-graph accelerators.

\begin{algorithm}[t]
\caption{Bit-plane schedules for one GEMM tile with $L$ weight and $R$ activation planes (plane-fetch counts in comments). The Temporal-Precision Engine is the activation-stationary schedule plus per-stage precision control; activations are re-fetched for every tile.}
\label{alg:engine}
\begin{algorithmic}[1]
\State \textbf{Naive bit-plane loop} \hfill $2LR$ fetches
\For{$i = 1$ \textbf{to} $L$}
    \For{$j = 1$ \textbf{to} $R$}
        \State fetch $W_i$;\ fetch $A_j$;\ \textsc{Exec}$(W_i, A_j,\ \mathrm{shift}{=}i{+}j)$
    \EndFor
\EndFor
\State
\State \textbf{Weight-stationary} \hfill $L + LR$ fetches
\For{$i = 1$ \textbf{to} $L$}
    \State fetch $W_i$ \Comment{held resident}
    \For{$j = 1$ \textbf{to} $R$}
        \State fetch $A_j$;\ \textsc{Exec}$(W_i, A_j,\ i{+}j)$
    \EndFor
\EndFor
\State
\State \textbf{Temporal-Precision Engine (ours)} \hfill $L_s + R$ fetches
\For{stage $s = 1$ \textbf{to} $S$}
    \State $L_s \leftarrow$ per-stage weight precision \Comment{one register write, $0$ cycles}
    \For{$j = 1$ \textbf{to} $R$}
        \State fetch $A_j$ \Comment{resident for the whole tile}
    \EndFor
    \For{$i = 1$ \textbf{to} $L_s$}
        \State fetch $W_i$ \Comment{streamed once}
        \For{$j = 1$ \textbf{to} $R$}
            \State \textsc{Exec}$(W_i, A_j,\ i{+}j)$
        \EndFor
    \EndFor
\EndFor
\end{algorithmic}
\end{algorithm}

\subsection{Hardware-Agnostic Design Principle}
\label{ssec:hw-principle}

TASQ represents adaptive precision independently of a particular kernel.
Each quantized layer stores one weight buffer at the maximum
bit-width $b_{\max}$ found across its temporal stages, plus a small bit-allocation map of
size $\mathcal{O}(L\cdot S)$ (number of layers $\times$ number of stages). For the
configurations evaluated in this paper, this map adds less than $0.005\%$ to the model's
weight footprint. At inference time, switching the effective precision of a layer is
implemented by reading fewer least-significant planes from the same buffer. It does not
require model reloading, kernel re-launch, or a runtime precision search.

The compiled kernel handles every stage and takes the number of bit planes as an argument;
the measured switch itself costs no cycles. Deployment then depends on the precision set
supported by the target hardware. The next section describes how we align TASQ's operating
precisions with that set.

\subsection{Adapting to Native Hardware Instructions}
\label{ssec:hw-native}

Given a target device with a set of natively supported integer precisions
$\mathcal{B}_{\text{native}} = \{b_1, b_2, \dots\}$, the user selects $b_{\max} =
\max\mathcal{B}_{\text{native}}$ as the storage precision and trains TASQ to operate over
$\mathcal{B}_{\text{native}}$. After training, every weight is physically stored at
$b_{\max}$ as a single $Q_{b_{\max}}$ buffer (Eq.~\eqref{eq:truncquant}), and per-stage
operating precision is realized at inference time by truncating its least significant
bits.

Generalizing the one-bit right-shift in Eq.~\eqref{eq:truncquant} to a
$k$-bit right-shift, the $b_j$-bit operand obtained from the stored $b_i$-bit code
($b_i > b_j$) is
\begin{equation}
Q_{b_j} \;=\; \left\lfloor \frac{Q_{b_i}}{2^{\,k_{i\to j}}} \right\rfloor,
\qquad k_{i\to j} \;=\; b_i - b_j,
\label{eq:hw_rightshift}
\end{equation}
so that $k_{i\to j}$ corresponds to the number of LSB planes dropped from the stored
buffer. The discarded $k_{i\to j}$-bit LSB block, generalizing the single-LSB
extraction in Eq.~\eqref{eq:lsb_extraction}, is
\begin{equation}
\mathrm{LSB}^{(k_{i\to j})} \;=\; Q_{b_i} \;-\; 2^{\,k_{i\to j}} \cdot Q_{b_j}
\;=\; \sum_{p=0}^{k_{i\to j}-1} 2^{\,p}\,Q_{b_i}[p],
\label{eq:hw_multi_lsb}
\end{equation}
where $Q_{b_i}[p] \in \{0,1\}$ denotes the $p$-th bit-plane of $Q_{b_i}$
(least-significant first). At inference, the trained mask $M_{t,l}$ has annealed to a
binary indicator that determines, per stage and per layer, which planes are kept;
the operating precision is then $b_j = b_{\max} - k_{i\to j}$, and
Eq.~\eqref{eq:hw_rightshift} is realized by reading only the top $b_j$ planes of the
stored buffer. No re-quantization, model swap, or per-stage checkpoint is needed.

Importantly, the same trained TASQ checkpoint can be re-deployed across backends with
different native precision sets simply by selecting a different operating precision
subset --- the stored weight buffer is unchanged. This decouples training from any specific deployment target, which
is the property that motivates our hardware-agnostic claim.

\subsection{Static-Graph Compilation Targets}
\label{ssec:hw-limit}

A class of mobile and edge accelerators, including the Apple Neural Engine
through Core ML, Qualcomm Hexagon, and Edge TPU, relies on ahead-of-time
compilation of a static computation graph in which the precision of each
layer is fixed at compile time. On these targets, the runtime cannot vary
the number of bit-planes consumed by a kernel call, because the graph
compiler has already lowered each op to a fixed-precision tensor
primitive. The same restriction applies to any method that changes layer precision at
runtime. Targets that dispatch work at run time rather than lowering it ahead of
time do not impose this constraint, since the number of bit-planes consumed by a
call can then be chosen per stage.

\section{Limitations and Future Directions}
\label{sec:appendix-limitations-future-work}

\paragraph{Commodity-hardware support.} The learned precision schedule can be evaluated at a matched operation budget on any platform, but cycle savings require hardware whose cost changes with operand precision~\citep{bismo,barvinn}. Datapaths whose lowest supported precision is four bits do not expose sub-4-bit execution, so a 2- or 3-bit TASQ layer still pays the four-bit cost. Supporting TASQ on emerging FP4 and microscaling formats will require production kernels that expose their finer precision choices.

TASQ targets the cost of each denoising step and can be used with faster samplers or distilled models; SDXL-Turbo shows that stage-wise allocation still helps on a four-step schedule. We leave prompt-conditioned masks and joint weight--activation precision to future work. Appendix~\ref{ssec:hw-limit} discusses runtime support.

\section{Additional Image Quality Results}
\label{sec:appendix-additional-image-quality-results}

\begin{figure*}[p]
    \centering
    \includegraphics[width=\linewidth,height=0.92\textheight,keepaspectratio]{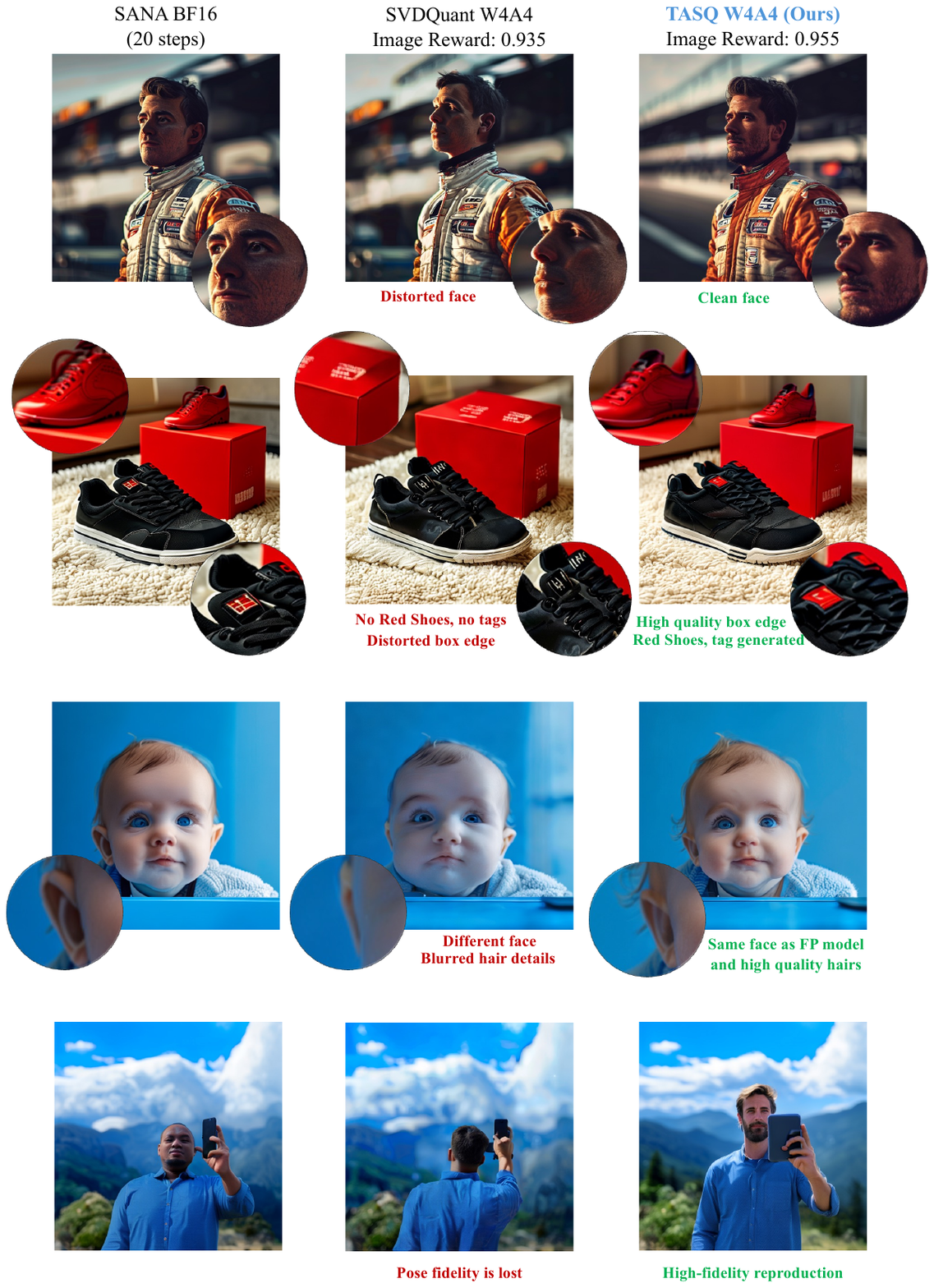}
    \caption{Qualitative Image Generation Results on SANA-1.6B}
    \label{fig:appendix_sample_sana}
\end{figure*}

\begin{figure*}[p]
    \centering
    \includegraphics[width=\linewidth,height=0.92\textheight,keepaspectratio]{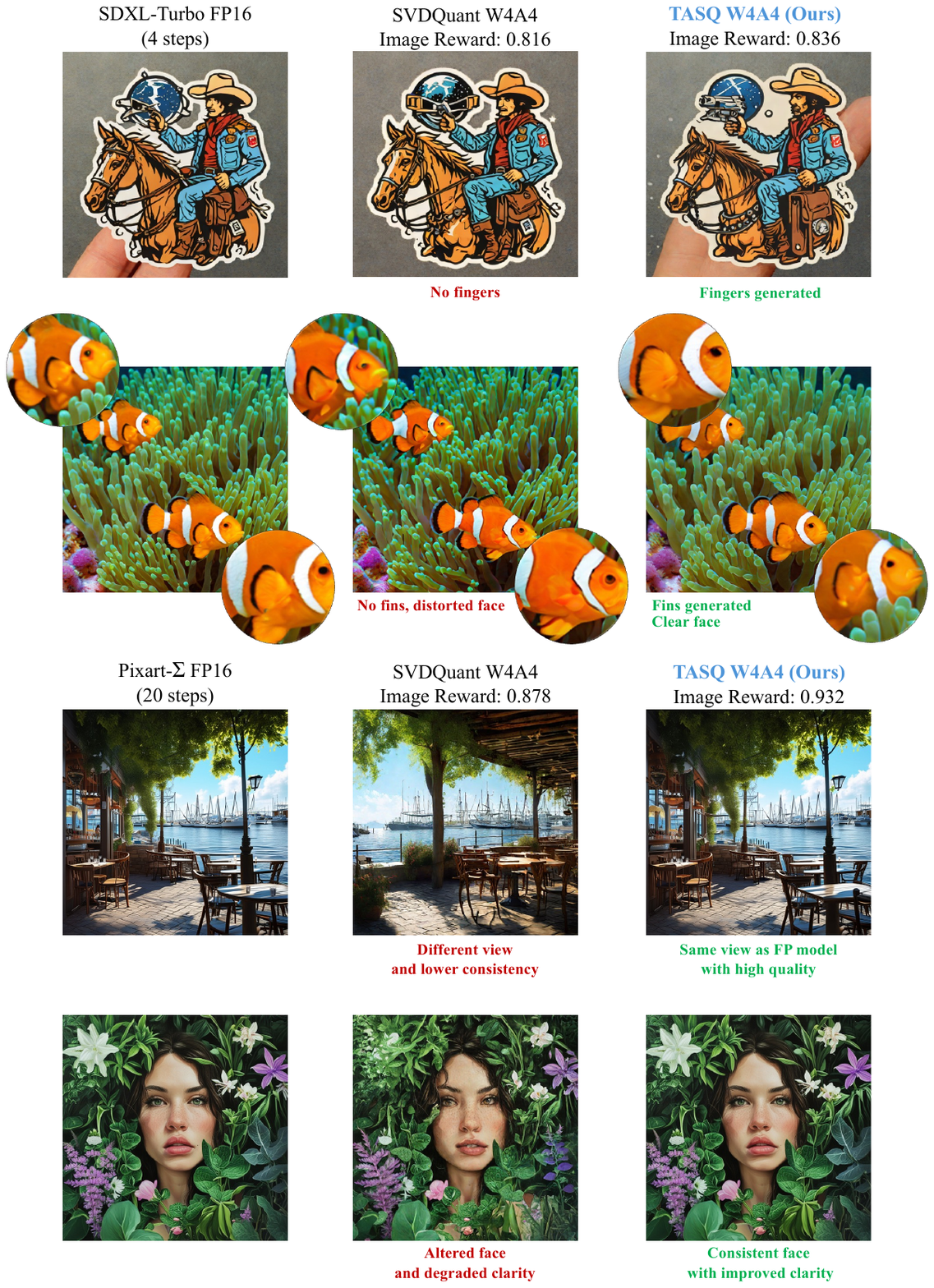}
    \caption{Qualitative Image Generation Results on SDXL-Turbo and PixArt-$\Sigma$}
    \label{fig:appendix_sample_sdxl_pixart}
\end{figure*}

Figures~\ref{fig:appendix_sample_sana} and~\ref{fig:appendix_sample_sdxl_pixart} provide additional comparisons with SVDQuant. In these examples, TASQ retains more of the objects and attributes present in the full-precision output, although both quantized models can differ visibly from that reference.

\end{document}